\documentclass[11pt, letterpaper]{arxiv}
\usepackage[all]{hypcap}
\usepackage[comma,numbers,sort,compress]{natbib}
\usepackage{hyperref}[citecolor=magenta]
\usepackage[capitalize,noabbrev]{cleveref}
\hypersetup{
    colorlinks = true,
    citecolor = {magenta},
}

\usepackage{microtype}
\usepackage{graphicx}
\usepackage{subfigure}
\usepackage{booktabs} 

\usepackage{xcolor}
\usepackage{colortbl}
\usepackage{algorithm}
\usepackage{algpseudocode}
\usepackage{setspace}

\usepackage{amsmath}
\usepackage{amssymb}
\usepackage{mathtools}
\usepackage{amsthm}

\usepackage{placeins}

\usepackage{fleqn, tabularx}

\usepackage{float}

\usepackage[most,skins,theorems]{tcolorbox}
\tcbset{
  aibox/.style={
    width=\linewidth,
    top=7pt,
    bottom=2pt,
    colback=rliableolive!13!white,
    colframe=black,
    colbacktitle=black,
    enhanced,
    center,
    attach boxed title to top left={yshift=-0.1in,xshift=0.15in},
    boxed title style={boxrule=0pt,colframe=white,},
  }
}
\newtcolorbox{AIbox}[2][]{aibox,title=#2,#1}

\usepackage{wrapfig}
\definecolor{lightblue}{rgb}{0.22,0.45,0.70}
\usepackage{fleqn, tabularx}

\definecolor{rliableolive}{HTML}{BBCC33}
\definecolor{rliableblue}{HTML}{77AADD}
\definecolor{rliablered}{HTML}{EE8866}

\usepackage{crossreftools}
\usepackage[suppress]{color-edits}

\usepackage{dsfont}
\usepackage{nicefrac}
\newtcolorbox{analysisbox}[1][]{
    enhanced jigsaw,
    colback=white,
    colframe=blue!75!black,
    fonttitle=\bfseries,
    boxsep=5pt,
    left=5pt,
    right=5pt,
    top=5pt,
    bottom=5pt,
    title=#1,
}
\definecolor{editInitialResponse}{RGB}{255, 235, 156} 
\definecolor{editBacktrack}{RGB}{0, 0, 139}
\definecolor{editRevisedResponse}{RGB}{255, 182, 193}

\usepackage{pifont}
\usepackage{soul}
\definecolor{highlightmistake}{RGB}{255, 179, 179} 
\definecolor{highlightcorrect}{RGB}{179, 255, 179}

\usepackage[capitalize,noabbrev]{cleveref}

\theoremstyle{plain}
\newtheorem{theorem}{Theorem}[section]

\theoremstyle{definition}
\newtheorem{definition}[theorem]{Definition}

\theoremstyle{remark}

\usepackage[textsize=tiny]{todonotes}
\usepackage{multirow}
\usepackage{enumitem}

\usepackage{amsmath,amsfonts,bm}

\def\eqref#1{Eq.~\ref{#1}}

\def\1{\bm{1}}

\DeclareMathAlphabet{\mathsfit}{\encodingdefault}{\sfdefault}{m}{sl}
\SetMathAlphabet{\mathsfit}{bold}{\encodingdefault}{\sfdefault}{bx}{n}

\newcommand{\bz}{\mathbf{z}}

\newcommand{\by}{\mathbf{y}}

\newcommand{\bx}{\mathbf{x}}

\newtcolorbox{promptbox}[2][]{  
listing only,
enhanced,
breakable,
colback=rliableolive!13!white,
colframe=black,
fontupper=\ttfamily,
title=#2,
#1}

\newcommand{\methodname}{\texttt{POPE}}
\newcommand{\methodfullname}{\textbf{Privileged On-Policy Exploration}}

\title{\texttt{POPE:} Learning to Reason on Hard Problems via Privileged On-Policy Exploration}

\author[1]{Yuxiao Qu$^*$}
\author[1]{Amrith Setlur$^*$}
\author[1]{Virginia Smith}
\author[1]{Ruslan Salakhutdinov}
\author[1]{Aviral Kumar}

\affil[1]{Carnegie Mellon University, ~~~($^*$Equal Contribution)}

\correspondingauthor{yuxiaoq@andrew.cmu.edu, asetlur@andrew.cmu.edu 
\newline
\textbf{Blog post:} \textnormal{\url{https://tinyurl.com/pope-blog}}; \textbf{GitHub:} \textnormal{\url{https://github.com/CMU-AIRe/POPE}}
}

\begin{document}

\maketitle

\begin{figure}[h]
\vspace{-0.5cm}
\centering
\includegraphics[width=0.85\linewidth]{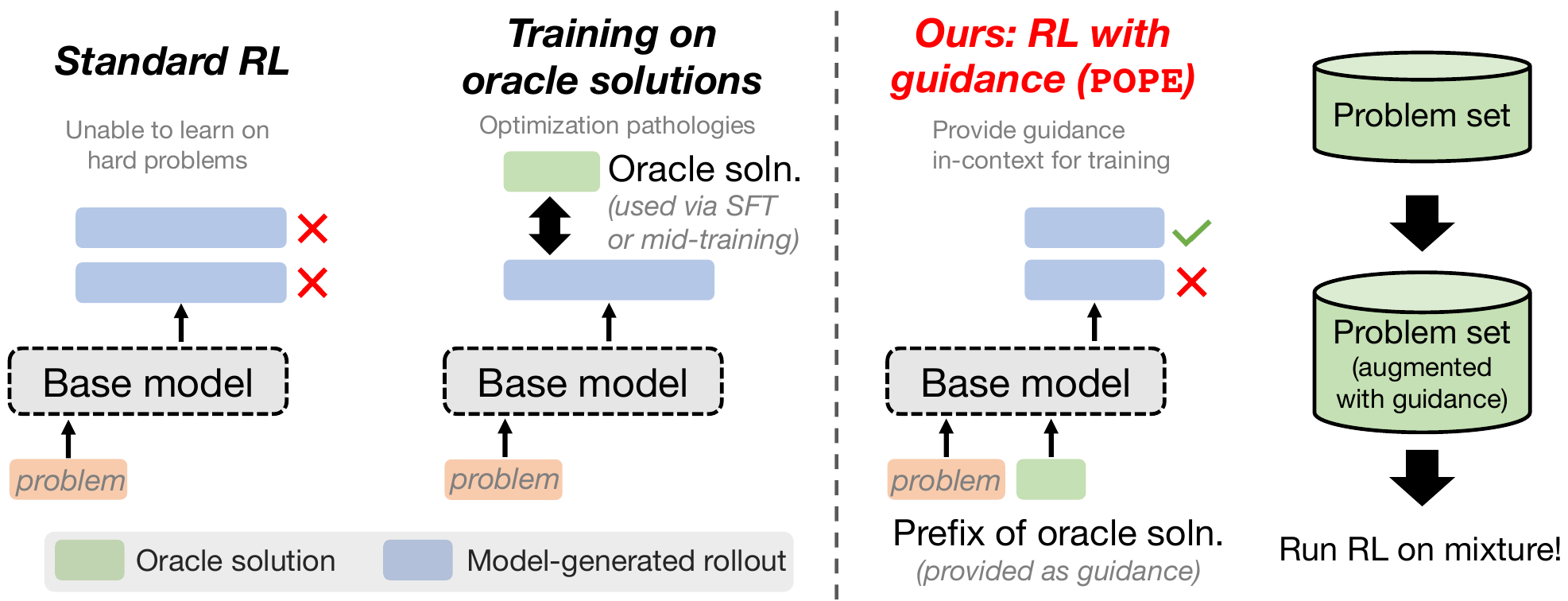}
\vspace{-0.2cm}
\caption{\footnotesize{\textbf{A schematic of our approach, \methodfullname{} (\methodname{}) compared with other approaches for training on hard problems,} where standard on-policy RL largely fails to produce successful rollouts. \methodname{} uses oracle (e.g., human-written) solutions to solely guide on-policy exploration during RL, without ever training on the oracle information as targets (for e.g., via supervised fine-tuning or mid-training). We show that modifying standard RL training objectives to incentivize token-level exploration frequently introduce optimization pathologies. Our training approach (\methodname{}) sidesteps such pathologies, while enabling learning from oracle information by ``instructing'' the model to build upon it.}}
\vspace{-0.2cm}
\label{fig:teaser}
\end{figure} 

{\absfont
\textbf{Abstract:} Reinforcement learning (RL) has improved the reasoning abilities of large language models (LLMs), yet state-of-the-art methods still fail to learn on many training problems. On \emph{hard} problems, on-policy RL rarely \emph{explores} even a single correct rollout, yielding zero reward and no learning signal for driving improvement. We find that natural solutions to remedy this exploration problem from classical RL, such as entropy bonuses, more permissive clipping of the importance ratio, or direct optimization of pass@k objectives, do not resolve this issue and often destabilize optimization without improving solvability. A natural alternative is to leverage transfer from easier problems. However, we show that mixing easy and hard problems during RL training is counterproductive due to \emph{ray interference}, where optimization focuses on already-solvable problems in a way that actively inhibits progress on harder ones. To address this challenge, we introduce \methodfullname{} (\methodname{}), an approach that leverages human- or other oracle solutions as \emph{privileged} information to guide exploration on hard problems, unlike methods that use oracle solutions as training targets (e.g., off-policy RL methods or warmstarting from SFT). \methodname{} augments hard problems with prefixes of oracle solutions, enabling RL to obtain non-zero rewards during guided rollouts. Crucially, the resulting behaviors transfer back to the original, unguided problems through a synergy between instruction-following and reasoning. Empirically, \methodname{} expands the set of solvable problems and substantially improves performance on challenging reasoning benchmarks.
}
\vspace{-0.3cm}
\section{Introduction}
\vspace{-0.2cm}
Reinforcement learning (RL) has significantly improved the reasoning abilities of large language models (LLMs) in domains such as math and coding. In particular, relatively small models trained with RL to better exploit test-time compute via longer chains of thought (CoT) can outperform much larger models trained without RL~\citep{setlur2025e3learningexploreenables,liu2025prorlprolongedreinforcementlearning}. While some works argue that RL post-training primarily amplifies existing capabilities~\citep{yue2025doesreinforcementlearningreally,zhao2025echochamberrlposttraining}, others show that careful design choices can mitigate these effects~\citep{setlur2025e3learningexploreenables,liu2025prorlprolongedreinforcementlearning}.
Across various RL recipes, a shared limitation is that on-policy RL fails to train on a large fraction of available problems, leaving substantial gains untapped. On-policy RL often cannot sample any non-zero–reward rollout on \emph{hard} problems relative to the base model, yielding no learning signal; for instance, when running \texttt{Qwen3-4B-Instruct} on DAPO-MATH-17K~\citep{yu2025dapo}, fewer than 50\% of problems produce a correct rollout even with $K=32$ attempts and a 16k token budget. Common throughput-oriented heuristics, such as dynamic sampling and zero-variance filtering, further discard these problems explicitly~\citep{yu2025dapo,wang2025reinforcementlearningreasoninglarge,khatri2025art}.

\textbf{How can we make progress on hard problems?} In a typical RL framing, this would require improving the ``exploration'' (i.e., rollout generation) mechanism used during learning. While standard on-policy RL relies on inherent stochasticity of the base model's distribution to guide exploration, on hard problems this na\"ive exploration strategy is insufficient. A natural attempt would be to employ token-level exploration bonuses from classical deep RL to incentivize exploration. We empirically analyze two representative methods from this category, and find that neither approach improves ``solvability'' (i.e., obtaining at least \emph{one} correct rollout when sampling multiple) without destabilizing optimization.

\begin{wrapfigure}{r}{0.25\textwidth}
\vspace{-0.5cm}
\centering
\includegraphics[width=0.97\linewidth]{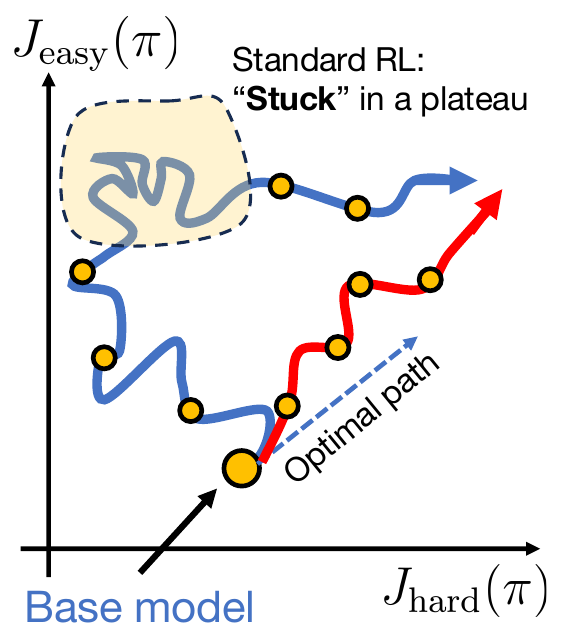}
\vspace{-0.2cm}
\caption{\footnotesize{\emph{\textbf{Interference~\citep{schaul2019ray}.}} 
In on-policy RL, training on a mixture of easy and hard problems preferentially accelerates progress on easy problems, often stalling or degrading performance on hard ones. This imbalance leads to plateaus during training; an ideal approach would induce more ``uniform'' progress across all problems.}}
\vspace{-0.4cm}
\label{fig:ray_int_teaser}
\end{wrapfigure} 
An alternative is to leverage \emph{transfer} to guide exploration. Useful skills learned on easy problem can then be chained to inform exploration on harder ones. To test whether such transfer enables exploration on hard problems, we train on mixtures of easy and hard problems. Through controlled experiments, we find that even when mixing in the most closely related easy problem, on-policy RL makes slow progress on a hard problem: it first ``sharpens'' the base model on the easy subset before improving on the hard one. We explain this behavior via \emph{\textbf{ray interference}}~\citep{schaul2019ray} (Figure~\ref{fig:ray_int_teaser}): an implicit bias in on-policy RL towards further optimizing reward on states where reward is already attained rather than finding reward on new states. Consequently, enabling learning on hard problems requires first obtaining non-zero reward by explicitly encouraging exploration some other way.

\textbf{If the base model cannot sample correct rollouts on a hard problem, (with high enough probability) how can we obtain non-zero reward?} A natural approach is to collect ``expert'' traces from humans/oracle and either distill them into the base model~\citep{sessa2024bondaligningllmsbestofn,agarwal2024onpolicy} or use them in RL as off-policy data~\citep{yan2025learningreasonoffpolicyguidance}. However, the type of reasoning traces that LLMs are trained to produce are prohibitively expensive to obtain, and prior work finds limited gains from available human-written data. Empirically, we find that distillation often caps gains from RL and off-policy training destabilizes RL. We therefore seek a more effective source of \emph{\textbf{exploratory signal}} on hard problems.

\underline{\textcolor{lightblue}{\emph{\textbf{Our key insight}}}} is that oracle solutions can effectively guide an LLM’s \emph{on-policy exploration} on hard problems, even when they are ineffective as training targets. Consider a hard problem where the LLM repeatedly follows incorrect approaches and fails within the training budget: conditioning on even a short prefix of a ``privileged'' human-written or oracle-provided solution can substantially increase the probability of reaching the correct answer. This effect is particularly pronounced when the base model has strong instruction-following capabilities, allowing us to steer it into building upon privileged content. \methodfullname{} (\methodname{}) leverages this principle to guide exploration in RL and this exploration is performed fully on-policy, providing an alternative to distillation or off-policy RL (Figure~\ref{fig:teaser}).

Concretely, for a set of hard problems, \methodname{} collects a human- or oracle-provided solution and uses a short prefix of this solution as privileged guidance during training. We train the base LLM with RL on a mixture of the original hard prompts and guided variants augmented with this fixed prefix (optionally together with easier prompts). Although these partial solutions are poor training targets, conditioning on them and ``instructing'' the policy to utilize them, reliably steers on-policy rollouts into regions where at least one correct attempt can be sampled. Behaviors learned under guidance through RL then transfer back to the original, unguided problems, greatly reducing difficulty of solving the hard problem from scratch. This transfer is enabled by \textbf{(a)} strong instruction-following, which allows the model to build on the prefix despite being unable to generate it itself, and \textbf{(b)} by backtracking and reflection behaviors that revisit and reinterpret the guidance during reasoning. 
When viewed through an RL lens, instruction-following and backtracking improves the overlap between the distribution of underlying states with and without any privileged guidance, which in turn enables transfer. Finally, from a classical RL perspective, \methodname{} mirrors a key principle from off-policy RL: learning from on-policy actions from off-policy states (problems + guidance) can be much more effective than learning from both off-policy states and actions~\citep{park2024value}.

\textbf{Results.} \methodname{} enables models to solve hard problems that remain unsolvable with standard RL, using either human-provided solutions. On a hard training subset, \methodname{} solves 10\% more problems measured via pass@16 with 64 rollouts and a 32k token budget. These gains persist even when training on mixtures of easy and hard problems, where guided exploration outperforms na\"ive mixtures and avoids collapse into sharpening on already-solvable problems. On standardized benchmarks such as AIME~2025 and HMMT~2025, \methodname{} consistently improves both pass@1 and pass@k, achieving up to 58\% pass@1 and 83\% pass@16 (vs. 48\% and 77\% for the base model), demonstrating robust population-level improvements.
\vspace{-0.2cm}
\section{Preliminaries and Notation}
\vspace{-0.2cm}

We study RL post-training of a base large language model (LLM), denoted by $\pi_\text{base}$ with parameters $\theta$. For any given input problem $\bx \sim \rho$ and a rollout $\by \sim \pi(\cdot \mid \bx)$ attempting to solve this problem, we define a \emph{binary outcome reward} $r(\bx, \by) \in \{0,1\}$ indicating correctness of the answer in these rollouts. Analogous to most work on RL with LLMs, we assume that the rollout $\by$ represents the final answer in a \texttt{\textbackslash boxed\{\}} block. We study several measures of performance, including the pass@$k$ metric, given by
\begin{align}
\label{eq:pass@k}
    [\text{pass@}k](\bx) = \Pr\!\big[\exists\, \by_1,...,\by_k \stackrel{\text { i.i.d. }}{\sim} \pi(\cdot \mid \bx) \;\; \text{s.t.}\;\; \max_{j=1}^k~~ r(\bx,\by_j) = 1\big],
\end{align} 
which measures the probability that at least one of $k$ independent attempts from a model $\pi$ at the problem $\bx$ succeeds. This metric captures the role of parallel exploration during training and measures whether a batch can yield any positive signal for policy gradient algorithms that do not train an explicit value function (e.g., GRPO~\citep{shao2024deepseekmathpushinglimitsmathematical}) and rely on Monte-Carlo rollouts for estimating the policy gradient. We use the empirical pass@$k$ value, denoted by $[\widehat{\text{pass@k}}](\bx)$ as a metric to quantify ``solvability'' of a problem $\bx$ during training. We estimate pass@$k$ by drawing $n$ independent samples from the model, with $n \gg k$.

\textbf{Outcome-reward on-policy RL.} Most RL algorithms train the base model $\pi$ with a policy gradient, which reinforces rollouts that end in a correct final answer, and reduces probability of rollouts that end up in the wrong answer (i.e., the negative gradient~\citep{setlur2025e3learningexploreenables}). This process is also called \emph{outcome-reward RL}. In practice, some of the most-commonly used RL algorithms such as GRPO, uses a reference policy $\pi_{\text{old}}$ for sampling, and normalize rewards into \emph{advantages} before utilizing them in the policy gradient: $A_i(\bx, \by_i) = r(\bx, \by_i) - \tfrac{1}{n}\!\sum_{j=1}^n r(\bx, \by_j)$, so that updates depend on deviations of reward from the batch mean. This normalized structure makes RL brittle on hard problems. If all $n$ rollouts fail on a given problem $\bx$ ($r(\bx, \by_i)=0$), then the advantage for all samples vanishes, $A_i=0$, and the gradient update is exactly zero on $\bx$. Thus when $[\widehat{\text{pass@}k}](\bx; \pi_\theta) \approx 0$, training stalls: advantages cannot generate signal, even with large batch sizes over problems or running training for longer. This creates a pathological feedback loop where the model sharpens on ``easy'' problems but halts learning on ``hard'' ones. 
\begin{tcolorbox}[colback=green!5!white,colframe=black,boxsep=0pt,top=4pt,bottom=4pt,left=3pt,right=3pt]
\begin{definition}[Hard and easy problems]
\label{def:hard_easy}
    A problem $\bx$ is called \emph{hard} for a given base model $\pi_\text{base}$ if for a sufficiently large value of $k$, $[\widehat{\text{pass@k}}](\bx; \pi_\text{base}) \approx 0$. A problem is called easy if it is not hard.
\end{definition}
\end{tcolorbox}
An important consideration when applying Definition~\ref{def:hard_easy} in practice is the output length used to evaluate this definition. Na\"ive empirical evaluations of the pass@k metric could underestimate its true value due to truncation of long model rollouts under low length budgets. This can make easy problems appear artificially harder. We find that training on such problems often does not pose a challenge, since models are able to ``compress'' their reasoning traces without any complex exploration problem. For our experiments, we therefore run all rollouts used to estimate $\widehat{\text{pass@k}}(\bx)$ until completion, up to 32k tokens for our base model \texttt{Qwen3-4B-Instruct}, and evaluate pass@k for values of $k$ up to 128.

\textbf{RL training loss.} Our approach and most of our analysis are both agnostic to the choice of the underlying training loss. But some of our analysis in Section~\ref{sec:rich_gets_richer} does utilize details of the RL training objective. We run a streaming, asynchronous implementation~\citep{piche2025pipelinerl} of GRPO~\citep{deepseekai2025deepseekr1incentivizingreasoningcapability,shao2024deepseekmathpushinglimitsmathematical} as our RL training algorithm, without any entropy and KL divergence terms as default. The GRPO loss uses a clipped surrogate similar to PPO~\citep{schulman2017ppo}, averaged over groups of trajectories. A typical loss function we optimize is:
\begin{align}
\label{eq:grpo}
\mathcal{L}_{\text{RL}}(\theta)
&= \mathbb{E}_{\mathbf{x},\mathbf{y}\sim\pi_{\rm old}}\left[
\min\left(
\frac{\pi_\theta(\mathbf{y}\mid \mathbf{x})}{\pi_{\rm old}(\mathbf{y}\mid \mathbf{x})}{A}(\mathbf{x},\mathbf{y}),
\;\text{clip}\!\left(\frac{\pi_\theta(\mathbf{y}\mid \mathbf{x})}{\pi_{\rm old}(\mathbf{y}\mid \mathbf{x})},\,1-\epsilon_\text{low},\,1+\epsilon_\text{high}\right){A}(\mathbf{x},\mathbf{y})
\right)
\right],
\end{align}
where ${A}(\mathbf{x}, \mathbf{y})$ denotes the advantage estimate discussed above, $\epsilon_\text{low}$ and $\epsilon_\text{high}$ are the low and high clipping thresholds. DAPO~\citep{yu2025dapo} sets $\epsilon_\text{high} > \epsilon_\text{low}$ to enable less conservative updates on positives that might be less likely under the sampling distribution $\pi_\text{old}$.

\vspace{-0.2cm}
\section{Why is Na\"ive Exploration Insufficient on Hard Problems?}
\label{sec:rich_gets_richer}
\vspace{-0.2cm}

To motivate the design of our approach in the next section, we first perform a systematic analysis to understand the efficacy and training dynamics of several exploration strategies when training on hard problems, where experiencing non-zero reward is challenging. The analogy to exploration in classical RL naturally suggests that seemingly straightforward techniques for encouraging exploration might help. 

We therefore study a representative subset of these techniques. The first type of methods perform ``token-level'' exploration by modifying the RL training objective or incorporating a bonus. The second type relies on transfer across problems by training on a mixture of easy and hard problems~\citep{sun2025rl}. In both cases, we observe characteristic failure modes pertaining to poor optimization or an amplification of ``interference'' where any strategies learned on easy problems do not guide learning on hard problems.

\vspace{-0.2cm}
\subsection{Token-Level Exploration on Hard Problems}
\label{sec:bonus_based}
\vspace{-0.2cm}
We first study the behavior of methods that incentivize token-level exploration on hard problems by training a \texttt{Qwen3-4B-Instruct} model on our hard problem set. Although this model cannot solve most of these problems initially, training for sufficiently many steps can yield non-zero reward on a small subset (approximately 6\%). To incentivize exploration, we experiment with two variants. 

\begin{figure}[t]
    \vspace{-0.3cm}
    \centering
    \includegraphics[width=0.95\linewidth]{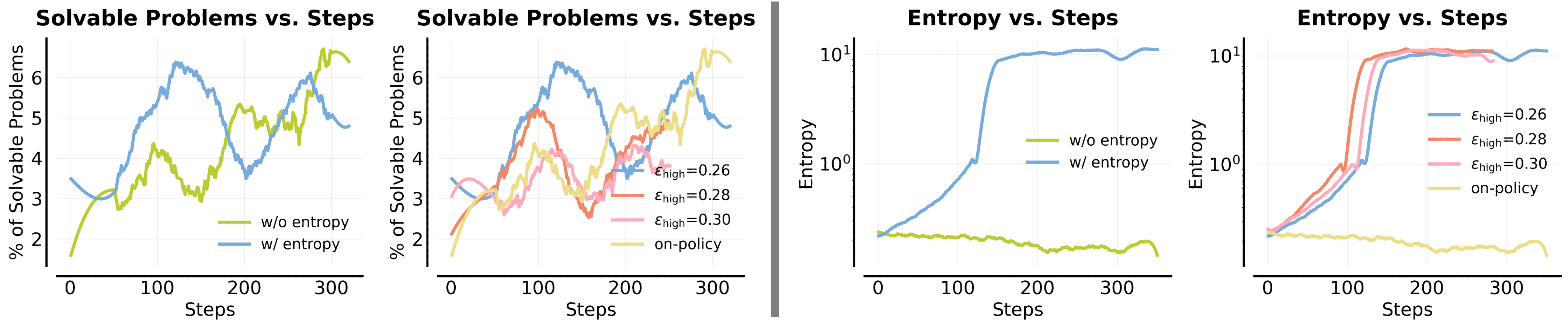}
    \vspace{-0.2cm}
      \caption{\footnotesize \textbf{Left: Evolution of the fraction of solvable problems} (measured via the pass@8 at 16k output length). \textbf{Right: average token-level entropy statistics over the course of RL training}. Observe that all of these representative classical exploration methods make similar amounts of (few) problems solvable, while creating pathologies in optimization in the sense that entropy blows up. We do notice large sensitivity to the clip threshold $\epsilon_{\text{high}}$ in our runs.}
    \label{fig:exploration_bonus}
    \vspace{-0.3cm}
\end{figure}

First, we add an entropy bonus together with a KL penalty to the policy objective. As shown in Figure~\ref{fig:exploration_bonus}, this modification does not make hard problems solvable: the fraction of problems with no correct solution among eight rollouts remains close to that of na\"ive RL throughout training. More problematically, the entropy of the model’s next-token distribution increases sharply, leading to an uncontrolled explosion early in training from which the model fails to recover.
Seeing this, we next attempt to increase exploration \emph{without} using an explicit entropy bonus. Specifically, we increase the high clip ratio, $\epsilon_\text{high}$ (Equation~\ref{eq:grpo}), in the update following DAPO~\citep{yu2025dapo}, with the goal of updating the model on rare positive traces that would otherwise be clipped. As shown in Figure~\ref{fig:exploration_bonus}, this approach also increases entropy, somewhat unexpectedly, but does not meaningfully improve solvability and performs no better than the entropy-based approach.

As discussed in Appendix~\ref{app:entropy_clip_ratio}, there is a systematic reason why the next-token entropy increases with a higher clip ratio. Briefly, when we use an importance-sampled policy gradient (Eq.~\ref{eq:grpo}) to train on rare positive traces, it attempts to shift probability mass toward these tokens using only a single gradient step. This both reduces confidence in tokens favored by the base model and fails to properly fit the positive trace, resulting in increased uncertainty and effectively random exploration. This increase in entropy snowballs over training resulting in entropy explosion. We detail this phenomenon in Appendix~\ref{app:entropy_clip_ratio}.

\begin{AIbox}{Takeaways: Token-level  exploration is insufficient on hard problems}
\begin{itemize}[itemsep=0pt]
 \setlength{\leftskip}{-15pt}
    \item Entropy bonuses cause uncontrolled entropy growth, inhibiting learning on hard problems.
    \item Higher clip ratios can help address the above but utilizing higher values results in amplification of entropy that ultimately results in random and meaningless exploration.
\end{itemize}
\end{AIbox}

\vspace{-0.2cm}
\subsection{Ray Interference Inhibits Exploration via Transfer}
\label{sec:ray_int}
\vspace{-0.2cm}

An alternative to token-level exploration is to leverage reasoning behaviors learned on easier problems as building blocks that can be composed to solve harder ones when given a larger token budget. This idea is referred to as \emph{extrapolation}~\citep{setlur2025e3learningexploreenables}: if training on easier problems produces a model that can use additional test-time compute to chain together multiple strategies, then on-policy RL may amplify this effect without needing specialized exploration  mechanisms.

\begin{figure}[ht]
    \vspace{-0.3cm}
    \centering
    \includegraphics[width=0.95\linewidth]{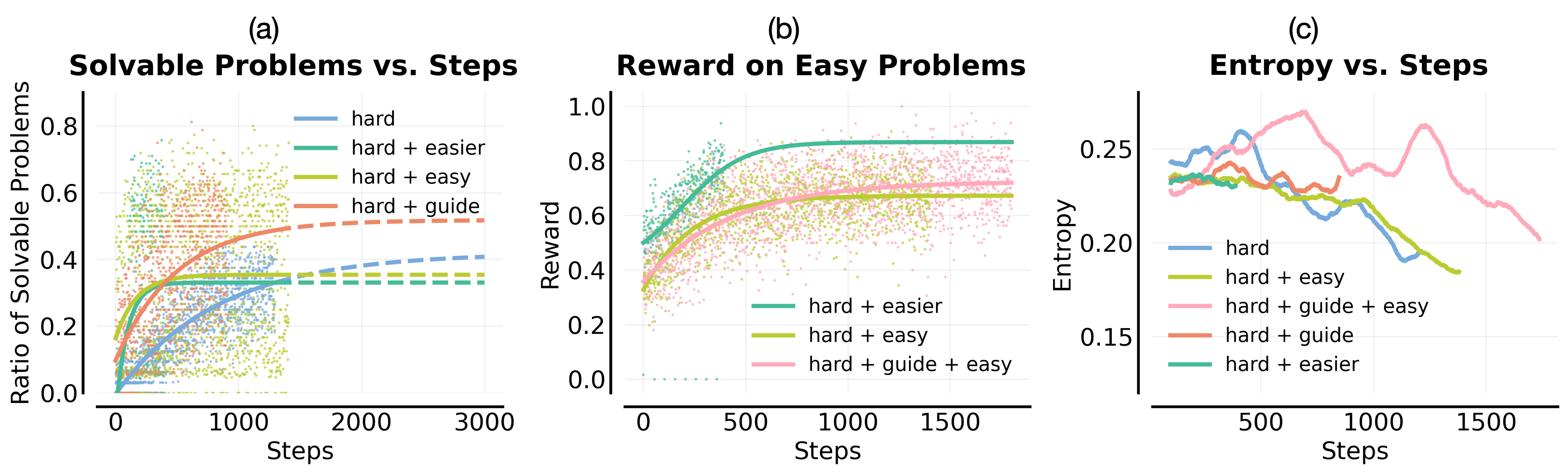}
    \vspace{-0.2cm}
      \caption{\footnotesize \textbf{No meaningful transfer from learning easy problems to hard problems.} \textbf{(a)} evolution of the fraction of solvable problems (measured via pass@8 at 16k response length). \textbf{(b)} average training reward on easy problems mixed in training. \textbf{(c)} average token-level entropy over the course of RL training. Since we do not use an entropy bonus, entropy generally remains stable (or slightly decreases) throughout training. Observe that the fraction of solvable problems increases the most when using our guidance-based approach, ``hard + guide''. In contrast, incorporating easy prompts does not improve solvability of hard problems, providing a negative result for the transfer hypothesis for improving exploration on hard problems.}
    \label{fig:easy_mix_zero_adv}
    \vspace{-0.3cm}
\end{figure}

To stress test whether transfer can guide exploration, we co-train on a mixture of easy and hard problems, with each subset containing 256 problems. Here, we define easy problems as those on which the base model achieves approximately 30\% success rate, evaluated with a 32k token budget and 128 rollouts, while easier problems correspond to those with roughly 60\% success rate under the same evaluation protocol. The motivation is that progress on easier problems during training might transfer to improved exploration and solvability on hard ones. As shown in Figure \ref{fig:easy_mix_zero_adv}, \emph{\textbf{we observe no meaningful improvement in solvability (pass@32) of hard problems}}. While mixing in easy problems (``hard + easy'') accelerates early gains in pass@32 on the hard set, the pass@32 performance quickly plateaus and converges to a lower asymptote than training on the hard problems alone. This indicates that learning on arbitrary easy problems does not transfer the exploration capabilities required to solve hard ones. A similar effect occurs when easier problems are mixed in (``hard + easier'' in Figure~\ref{fig:easy_mix_zero_adv}), which in fact results in even fewer hard problems being solved during training. In contrast, our approach (that we discuss in the next section) yields higher solvability rates, improving pass@8 by approximately 13\% relative to all mixture-based baselines. These results show that transfer from easy problems is insufficient for exploration.

\begin{figure}[t]
    \centering
    \vspace{-0.1cm}
    \includegraphics[width=0.99\linewidth]{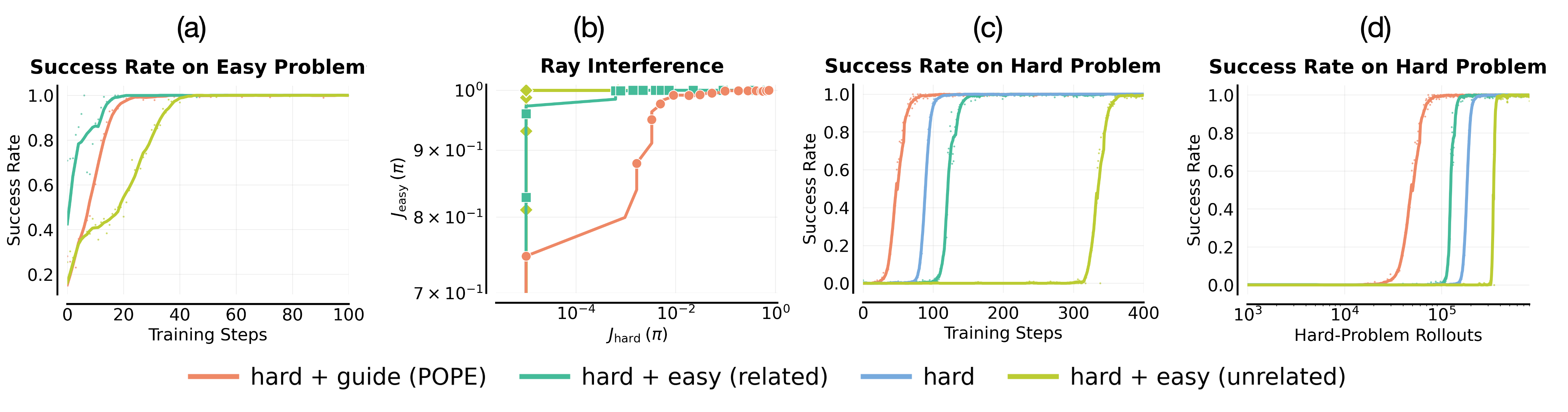}
    \vspace{-0.2cm}
      \caption{\footnotesize \textbf{Didactic two-problem experiment illustrating ray interference.} We train on a setting consisting of one easy and one hard problem. 
\textbf{(a)} Success rate on the easy problem versus training steps. All methods rapidly solve the easy problem. \textbf{(b)} Optimization trajectories visualized by plotting $J(\pi_\theta;\text{easy})$ and $J(\pi_\theta;\text{hard})$ jointly over training. Mixing in an unrelated easy problem leads to rapid improvement on the easy problem at the cost of stagnation on the hard problem, illustrating negative transfer due to ray interference. 
\textbf{(c)} Using a related easy problem partially mitigates this effect, but remains inefficient and requires many more training steps to solve the hard problem compared to training on the hard problem alone. Our approach (``hard + guide'') is the only one that improves convergence speed on the hard problem of all methods.
\textbf{(d)} Success rate on the hard problem vs. the number of rollouts allocated to it. Beyond interference, \methodname{} improves sample efficiency by reducing the number of rollouts required to learn the hard problem, indicating an acceleration in solvability of the hard problem.
}
    \label{fig:one_problem}
    \vspace{-0.4cm}
\end{figure}

\textcolor{lightblue}{\textbf{\emph{Didactic experiment with only one easy and one hard problem.}}} To conceptually understand why training on a mixture of easy and hard problems does not help, we run RL training in a didactic setting consisting of only \emph{two} problems: one easy and one hard and show results in Figure~\ref{fig:one_problem}. As expected, training on only the hard problem (``hard'' in Figure~\ref{fig:one_problem}c and \ref{fig:one_problem}d) often yields zero reward until the model succeeds once due to randomness, after which learning picks up and reinforces this success pretty quickly. Mixing in a very related easy problem, as measured by cosine similarity between the textual embeddings of the hard and easy problems under the base model, slightly accelerates training; see ``hard + easy (related)'' in Figure~\ref{fig:one_problem}. In contrast, mixing in an easy but unrelated problem that exhibits the lowest cosine similarity with the hard problem (``hard + easy (unrelated)'' in Figure~\ref{fig:one_problem}) substantially slows convergence on the hard problem to a point where RL learns to solve the hard problem much slower than simply training on the hard problem alone. This is a form of \textbf{\emph{interference}} between learning on different problems. Crucially, the related and unrelated easy problems (as well as the guided variant) are \emph{matched in base difficulty}: under the base model, they exhibit similar success rates.

We further visualize the optimization trajectory by plotting rewards on the easy and hard problems, $J(\pi_\theta;\text{easy})$ and $J(\pi_\theta;\text{hard})$, against each other over training in Figure~\ref{fig:one_problem}b. Across all settings, the easy problem begins accumulating reward early. When the easy problem is unrelated, RL preferentially optimizes $J(\pi_\theta;\text{easy})$ while progress on $J(\pi_\theta;\text{hard})$ stagnates, a form of negative interference consistent with \textbf{\emph{ray interference}}~\citep{schaul2019ray}. Ray interference is fundamentally a function-approximation effect in on-policy RL: the same mechanisms that enable transfer across related tasks can hinder learning when problems are semantically disjoint or exhibit large performance skew. Although related easy problems partially mitigate this effect, optimization on the hard problem remains slow. In contrast, applying \methodname{} enables smoother optimization of $J(\pi_\theta;\text{hard})$ (“hard + guide (\methodname{})” in Figure~\ref{fig:one_problem}b), yielding a more favorable trajectory that reduces interference from the easy prompt and improves exploration. This experiment isolates ray interference in a minimal setting and explains why na\"ive transfer from easy to hard is insufficient.

\textcolor{lightblue}{\textbf{\emph{Can we solve this issue by optimizing the empirical pass@k metric directly?}}} A natural next question is whether directly optimizing the empirical pass@k objective can address the lack of progress on hard problems. Prior work has proposed optimizing pass@k-style rewards to encourage population-level diversity and reduce sharpening~\citep{chow2024inference,walder2025pass}. While this approach can mitigate distribution collapse on problems where the model already attains occasional successes, it does not resolve the core difficulty on hard problems where the pass@1 score attained by the base model are quite low.

\begin{figure}[t]
    \vspace{-0.3cm}
    \centering
    \includegraphics[width=0.9\linewidth]{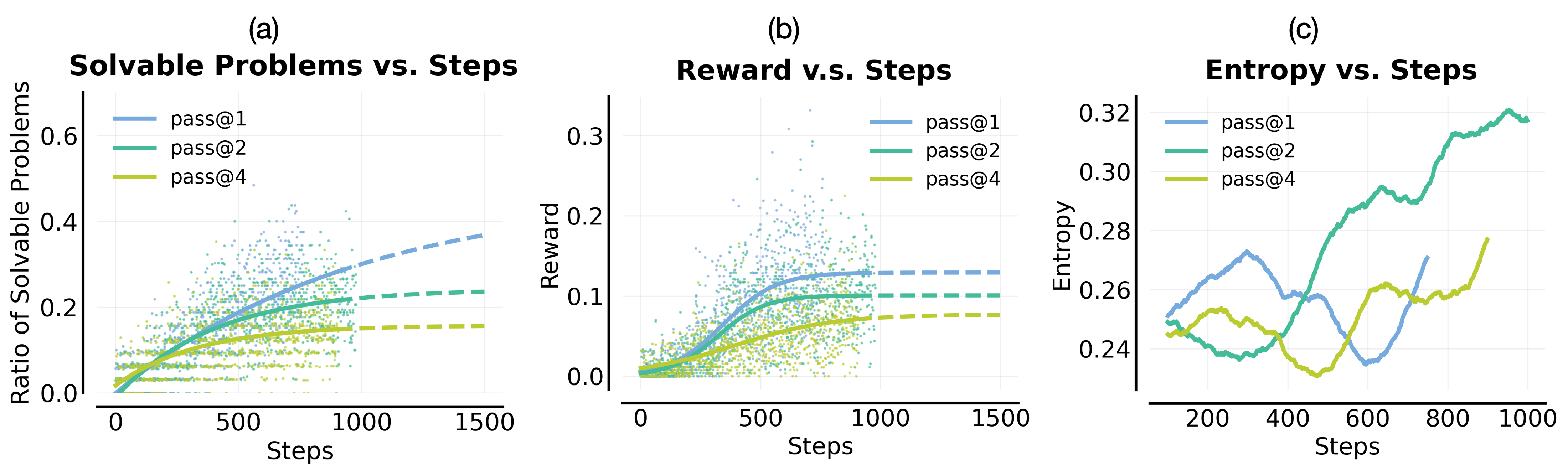}
    \vspace{-0.2cm}
      \caption{\footnotesize \textbf{Directly optimizing pass@k fails to improve exploration on hard problems and primarily prevents over-sharpening on already-solvable ones.}
\textbf{(a)} Evolution of the fraction of solvable hard problems under different pass@k objectives (measured at pass@8).
\textbf{(b)} Average training reward when optimizing pass@k compared to standard on-policy RL (pass@1).
\textbf{(c)} Average token-level entropy during training.
Although pass@k optimization is intended to promote population-level diversity, it does not improve hard-problem solvability. Instead, increasing $k$ consistently degrades performance relative to pass@1. These results indicate that pass@k optimization cannot bootstrap learning when the initial success probability is near zero: it primarily redistributes reward to incorrect traces to reduce over-sharpening, reinforcing already-solvable problems rather than enabling exploration on previously unsolved ones.}
    \label{fig:pass@k_optimization}
    \vspace{-0.3cm}
\end{figure}
As shown in Figure~\ref{fig:pass@k_optimization}, increasing $k$ consistently degrades performance, both in pass@$k$ (solvability) and in average pass@1 reward. While a drop in pass@1 is expected when optimizing pass@$k$ due to an objective shift, we find that pass@$k$ optimization also fails to improve solvability.
Why? Consider a setting in which hard problems are independent, so reward obtained on one problem does not transfer to others. In this regime, pass@$k$ optimization can improve solvability only if the model achieves non-zero pass@1 on each problem (since pass@$k$ is a monotonic function of per-problem pass@1), which does not hold for hard problems. Moreover, even when correct rollouts occasionally exist, pass@$k$ optimization redistributes reward toward incorrect traces to encourage diversity, shrinking the reward gap between positive and negative samples. On hard problems, where correct rollouts are already hard to sample, this inhibits learning. As a result, pass@$k$ optimization may mitigate over-sharpening but is ineffective for driving exploration and can slow convergence. We provide a detailed analysis of this behavior, including the pass@$k$ objective and its policy-gradient estimator, in Appendix~\ref{app:pass_at_k_more}.

\begin{AIbox}{Takeaways: Ray interference hurts exploration on hard problems}
\begin{itemize}[itemsep=0pt]
 \setlength{\leftskip}{-15pt}
    \item As RL starts solving some easy problems, its ability to solve other hard problems reduces.
    \item This phenomenon can be explained via ray interference from multi-task RL. Ray interference leads to stagnation and inefficient performance improvement on hard problems.
\end{itemize}
\end{AIbox}

\vspace{-0.25cm}
\section{\texttt{POPE}: Privileged On-Policy Exploration}
\label{sec:pope}
\vspace{-0.2cm}

In this section, our goal is to develop an exploration approach that enables the model to learn how to solve new, hard problems. To address the limitations of pure on-policy exploration, we leverage oracle solutions, such as human-written solutions, during training. A natural approach would be to train directly on these oracle solutions, either by imitating them via supervised fine-tuning (SFT) before running standard on-policy RL, or by incorporating them directly as additional rollouts during RL. However, we find that both approaches distort the base model’s reasoning patterns and lead to optimization instabilities.

\textbf{Limitations of training on oracle solutions.} Concretely, running SFT on human-written solutions over multiple epochs to closely fit the oracle data causes the model to memorize these solutions, resulting in a low-entropy initialization that inhibits meaningful exploration during subsequent RL. Conversely, early stopping SFT to avoid memorization yields a high-entropy initialization that cannot reliably produce rollouts in the style of either the base model or the oracle solutions. In both cases, the resulting policy is not effective enough for further improvement or generalization. We compare against improved variants of SFT and off-policy RL in our experiments in Section~\ref{sec:exps}. Next we develop our approach.

\begin{figure}[t]
    \centering
    \vspace{-0.4cm}
    \includegraphics[width=0.9
    \linewidth]{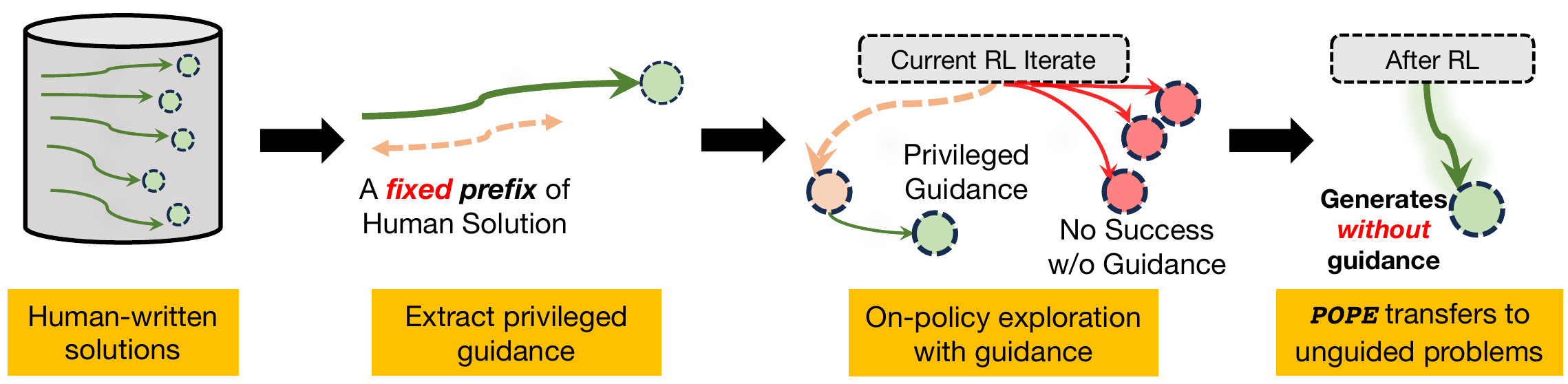}
    \vspace{-0.2cm}
      \caption{\footnotesize \textbf{Illustration of our approach \methodname{}.} \methodname{} trains the model by using privileged guidance from human solutions to condition on-policy generations. We show that training on a mixture of guided and unguided problems then allows transfer of the learned reasoning strategies to the unguided problem.}
    \label{fig:pope_main_fig}
    \vspace{-0.5cm}
\end{figure}

\textbf{Our approach.} Rather than using oracle solutions as training targets, \textcolor{lightblue}{\textbf{our key idea}} is to use them solely to \emph{steer} on-policy rollouts. We augment each hard problem with guidance in the form of a short prefix of an oracle solution and instruct the model to follow and build upon this guidance (see the system instruction at the end of this section). Although the model cannot generate such sequences on its own, conditioning on the partial solution moves it into more favorable regions of the response space from which non-zero reward becomes attainable.
From an RL perspective, this corresponds to initializing rollouts from off-policy ``states'' informed by human-written solutions, while learning remains fully on-policy. We train on a mixture of unguided hard problems and their guided variants, optionally including easy problems to broaden coverage. We find that this mixture enables behavior learned under guidance to transfer to unguided hard problems (Section~\ref{sec:why_does_pope_work}), often mitigating ray interference and improving overall success. We refer to this approach as \textbf{privileged on-policy exploration} (\methodname{}; Figure~\ref{fig:pope_main_fig}).

\textbf{Formal description.} Formally, given an oracle solution $\bz$ to a hard training problem $\bx \sim \mathcal{D}_\text{hard}$, we generate rollouts conditioned on a prefix $\bz^{0:i}$ of $\bz$ and a system instruction $I$ that ``instructs'' the model to build upon $\bz^{0:i}$, i.e., $\by \sim \pi(\cdot|\bx, \bz^{0:i}, I)$. In principle, any prefix $\bz^{0:i}$ could be used as guidance. However, an overly long prefix that solves a substantial portion of the hard problem is not useful for learning reasoning strategies that can transfer to the unguided setting. We therefore restrict ourselves to a short prefix that is sufficient to enable on-policy rollouts to obtain \emph{some} non-zero reward on $\bx$. To identify such a prefix, we evaluate the base model’s ability to produce at least one successful rollout when conditioned on a set of coarsely chosen, uniformly spaced prefixes, and select the shortest prefix that yields a successful trace under the base model. 
Let's denote $i^*(\bx)$ as the length of this short prefix for a problem $\bx$.
On problems where no prefix leads to a successful rollout, we simply utilize a randomly-chosen prefix that is smaller than $\nicefrac{1}{4}^{\mathrm{rd}}$ of the oracle solution. Using this, we construct a guided set of hard problems:
\begin{align}
    \label{eq:augmented_set}
    \mathcal{D}^\text{guided}_{\text{hard}} := \left\{ \text{concat}(\bx, \bz^{0:i^*(\bx)}, I) ~|~ \bx \in \mathcal{D}_\text{hard} \right\}.
\end{align}
\methodname{} then trains on a dataset consisting of a 1:1 mixture of (unguided) hard problems $\mathcal{D}_{\text{hard}}$ and their guided versions $\mathcal{D}^\text{guided}_{\text{hard}}$. Finally, we emphasize that \methodname{} operates fully on-policy: although privileged information guides exploration, the exploration itself is carried out by the model via on-policy rollouts.

\begin{tcolorbox}[
    title=\textbf{\methodname{} System Instruction},
  colback=blue!5,
  colframe=black,
  boxrule=0.6pt,
  width=0.99\linewidth,
  boxrule=1.0pt,
  arc=2pt,
  left=4pt,
  right=4pt,
  top=4pt,
  bottom=4pt]
\begin{minipage}{0.95\linewidth}
\small
You are given a problem and a partial solution. Your task is to \textbf{carefully study the partial response, identify what reasoning or steps are already provided, and then complete the solution from where it left off}. Ensure your response is logically consistent and leads to a complete and correct final answer.

~~

\textbf{Important}: \textit{Show your reasoning step-by-step, and present the final answer using LaTeX-style \(\boxed{\cdot}\).}

Problem: <Problem>

Partial Response: <Partial Response>

Continue solving the problem, starting from where the partial response ends. Make sure your final answer is written as: \(\boxed{\text{your answer here}}\)

\end{minipage}
\end{tcolorbox}

We find that \methodname{} enables models to gradually learn to solve unguided versions of hard problems that standard RL on the base model fails to solve, resulting in a form of transfer that we analyze next.

\begin{AIbox}{Summary: Privileged On-Policy Exploration (\methodname)}
\begin{itemize}[itemsep=2pt]
 \setlength{\leftskip}{-15pt}
    \item \methodname{} conditions on partial solutions from an oracle as privileged information to guide on-policy rollouts during RL training, instead of directly using oracle data as training targets.
    \item We identify a short prefix of the oracle solution that enables the base model to succeed once to augment a hard problem. We train on a 1:1 mixture of guided and unguided problems.
\end{itemize}
\end{AIbox}

\vspace{-0.2cm}
\section{Why Does \methodname{} Work?}
\label{sec:why_does_pope_work}
\vspace{-0.2cm}

We now conceptually and empirically study why learning on guided versions of hard problems transfers to improving performance on their unguided counterparts when training with \methodname{}. Since the model is never trained to imitate the guidance tokens themselves, the source of this transfer is not immediately obvious. Our explanation is based on a simple mental model in which \emph{stitching} plays a central role.

\vspace{-0.3cm}
\subsection{A Mental Model}
\vspace{-0.2cm}

To build intuition for why \methodname{} works, we consider a simple mental model of exploration in a Markov decision process (MDP). Suppose that obtaining reward from the initial state requires extensive exploration, but that there exists an intermediate subset of states, denoted $\mathcal{S}_\text{good}$, from which reward can be obtained reliably via standard on-policy sampling. Early in training, the agent is unaware of these states, as it has not yet experienced any reward. Guidance acts as a roll-in policy that steers the agent into $\mathcal{S}_\text{good}$, where learning signal becomes available and RL can proceed. On-policy RL from these states then learns an effective continuation policy in a region of the state space where reward is attainable.

Once such continuations are learned, the unguided policy no longer requires guidance to succeed from $\mathcal{S}_\text{good}$; it only needs to reach these states through its own behavior. Crucially, identifying whether a state belongs to $\mathcal{S}_\text{good}$ is itself difficult without evidence of success from that state. Training with guidance creates this evidence by learning successful completions conditioned on reaching $\mathcal{S}_\text{good}$. As a result, obtaining positive-reward traces from the initial state reduces to reaching some $s' \in \mathcal{S}_\text{good}$, after which the learned policy can already succeed. Once such traces are available, training further reinforces the behavior that leads to $s'$ from the initial state. In contrast, unguided RL must discover both $\mathcal{S}_\text{good}$ and the successful behaviors from scratch, making exploration significantly more challenging.

\begin{wrapfigure}{r}{0.7\textwidth}
\vspace{-0.3cm}
\centering
\includegraphics[width=0.97\linewidth]{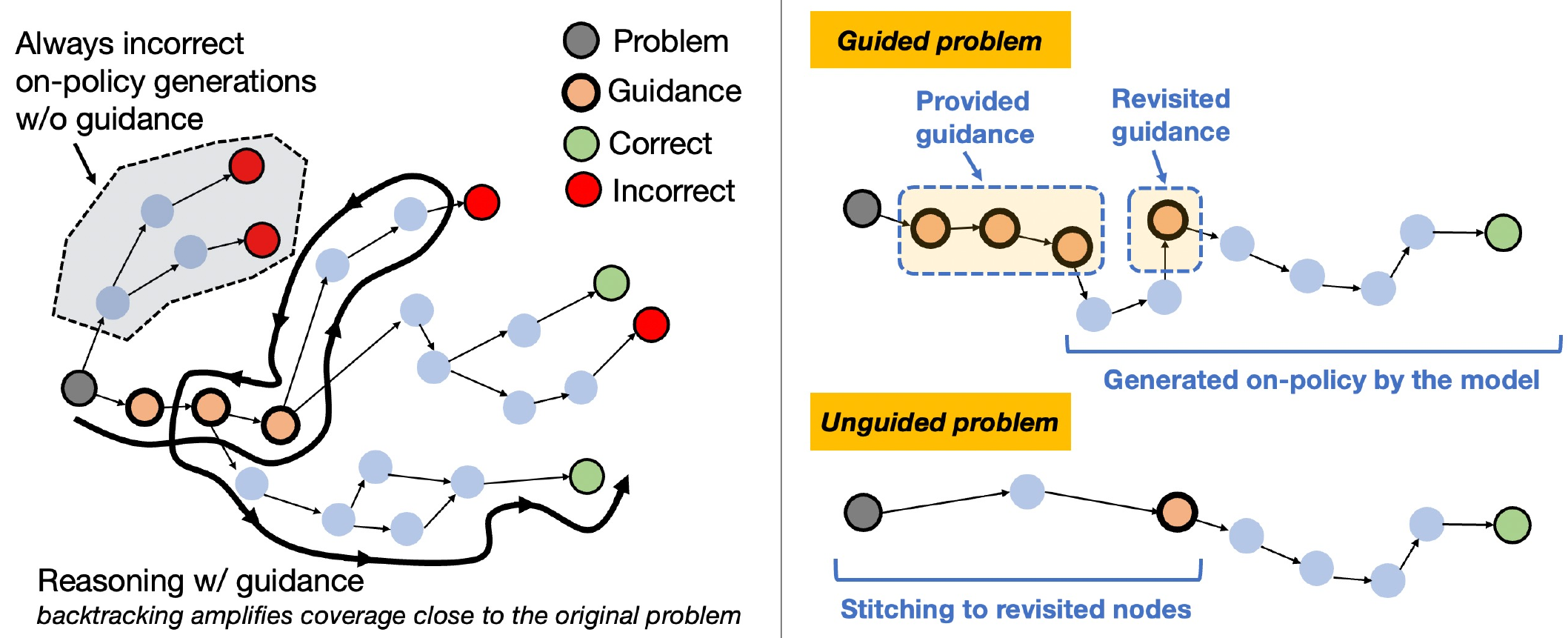}
\vspace{-0.2cm}
\caption{\footnotesize{\textbf{Illustration of how reasoning structure and instruction following enable performance improvements.} Instruction-following capabilities of the base model enable the policy to pursue reasoning paths (shown as a thick black line) that reach regions of the solution space where reward can be attained. Self-verification and backtracking behaviors in reasoning traces then allow the LLM to revisit states close to the initialization and construct successful continuations from there, amplifying coverage over states near the initial problem from which success is possible. By doing so, \methodname{} reduces the challenge of attaining reward on the original unguided problem to reaching a nearby state from which successful rollouts have already been experienced on the guided problem.}}
\vspace{-0.3cm}
\label{fig:mental_model}
\end{wrapfigure} 
\textbf{Applying this mental model to LLMs.} We now apply this mental model to LLMs. In an autoregressive MDP, a natural notion of state is the entire sequence of tokens produced so far. However, reasoning traces often exhibit substantial redundancy, suggesting that a more accurate notion of state for reasoning is the internal representation induced by a partial sequence, where newly generated segments can overwrite or revise earlier computation or attempts, resulting in revisiting similar states multiple times during a rollout. Guidance steers the model into internal states from which successful completions are more likely. The efficacy of this steering depends on whether the base model can follow the system instruction to build upon the guidance and comprehend the information it contains, even when the guidance itself consists of tokens that are unlikely under the base model. Models with strong \textbf{\emph{instruction-following}} capabilities can benefit from this mechanism, and obtain non-trivial reward signal on guided versions of hard problems.

Once the model has learned to solve the problem from states reached under guidance, the remaining challenge is to stitch these behaviors with those from the initial state. In general, it is unclear whether the base model would ever sample traces that perform computations similar to the provided guidance, especially when the guidance required to obtain a successful completion is long. In the MDP terminology above, the set of states $\mathcal{S}_\text{good}$ may itself be difficult to reach.
In this regime, the structure of \emph{\textbf{reasoning}} traces in long chain-of-thought models plays a central role in reducing the effective difficulty of reaching $\mathcal{S}_\text{good}$. Such models often self-verify, revisit earlier steps, and backtrack during generation (Figure~\ref{fig:mental_model}, left). When these behaviors occur in guided rollouts, they expand the model’s coverage over states closer to the problem that are otherwise unlikely to be sampled under base model rollouts without guidance. As a result, RL training begins to observe reward not only from the guided states induced by the oracle prefix, but also from a neighborhood of states that the model can plausibly reach without guidance. 

As a result, learning on the unguided problem becomes plausible. Rather than discovering reward from scratch, the policy only needs to explore to reach nearby states that were already visited during guided rollouts and the structure of reasoning amplifies \textbf{\emph{overlap}} under function approximation (Figure~\ref{fig:mental_model}, right). \emph{{This explains why \methodname{} enables transfer from guided to unguided problems}}. See Appendix~\ref{app:extended_overlap} for an extended discussion and further intuition about the overlap mechanism. Notably, we find that this transfer occurs even when \emph{\textbf{a fixed segment of guidance}} is used for training, despite conventional wisdom of coverage~\cite{chang2024dataset} demanding more segments for a transfer of performance to the unguided version.

\vspace{-0.25cm}
\subsection{Empirically Validating the Stitching and Overlap Hypothesis}
\label{sec:empirical_overlap}
\vspace{-0.2cm}

As discussed above, despite using a fixed guidance segment, \methodname{} enables transfer because backtracking and revision behaviors expand coverage over nearby states that an unguided rollout can plausibly reach. We now test our model above via an intervention that selectively reduces overlap between guided and unguided rollouts. If this overlap is important, then discouraging the model from revisiting earlier parts of the guided solution should weaken transfer from guided to unguided problems.

\textbf{Experimental setup.}
We modify the system instruction (shown below) in \methodname{} instructing the model to continue solving the problem in a guided rollout, \emph{without} restating, paraphrasing, or recomputing any part of the guidance. This instruction encourages the model to treat the guidance as a silent scaffold and to avoid backtracking to intermediate steps that would otherwise be revisited. 

\begin{tcolorbox}[
title=\textbf{Modified \methodname{} System Instruction},
  colback=blue!5,
  colframe=black,
  boxrule=0.6pt,
  width=0.99\linewidth,
  boxrule=1.0pt,
  arc=2pt,
  left=4pt,
  right=4pt,
  top=4pt,
  bottom=4pt]
\begin{minipage}{0.95\linewidth}
\small
You are given a problem and a partial solution. Your task is to infer what reasoning has already been completed and continue solving the problem \textbf{without repeating, paraphrasing, or referencing any part of the partial response}. You must \textbf{not} restate earlier steps, summarize them, or quote them in any form. Begin directly from the next logical step that has not yet been completed.

~~

\textbf{Important:} \textit{Use the information from the partial response silently. Do not copy, rephrase, or explicitly mention anything from it. Your continuation must be logically consistent with what has already been done. Show your reasoning step by step (only the new steps), and present the final answer using \(\boxed{\cdot}\) notation.}
\end{minipage}
\end{tcolorbox}

\begin{wrapfigure}{r}{0.72\textwidth}
\vspace{-0.5cm}
\centering
~~~~\includegraphics[width=0.99\linewidth]{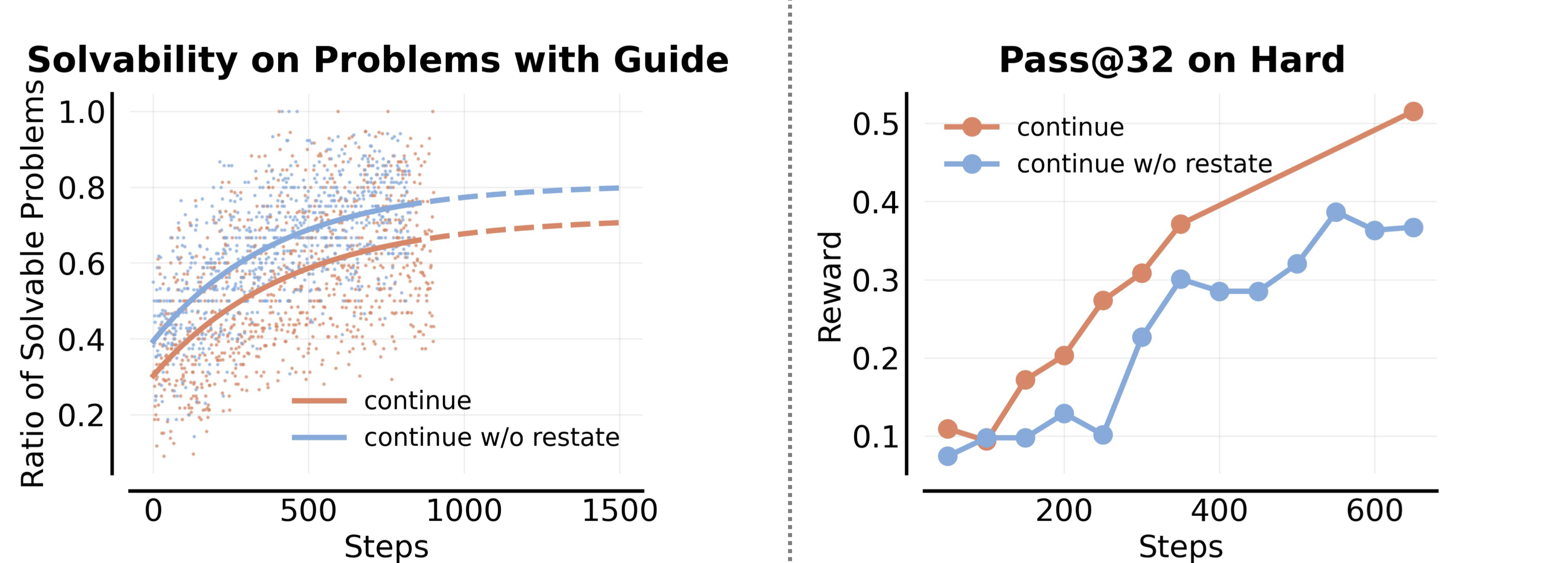}
\vspace{-0.6cm}
\caption{\footnotesize{\textbf{Left: solvability (pass@8)} and \textbf{Right: pass@32 scores on the guided and unguided versions of the hard prompt}. The system instruction that forces the model to continue without restating or revisiting information in the guidance solves more problems with guidance, presumably because it simplifies the RL problem conditioned on the guidance. However, this system instruction also achieves a worse pass@32 score on the unguided version of the hard problem, indicating reduced transfer from guided to unguided settings, supporting our mental model.}}
\vspace{-0.3cm}
\label{fig:prompt_study}
\end{wrapfigure}

\textbf{Results.}
As shown in Figure~\ref{fig:prompt_study}, this modification to the system instruction shifts performance in a manner consistent with the stitching/overlap mental model. The modified instruction improves performance on the \emph{guided} version of the hard problems, consistent with making the RL problem easier conditional on guidance. However, it reduces transfer to the \emph{unguided} problems, yielding a lower pass@32 score compared to the default instruction used in \methodname{}. In effect, the intervention biases learning toward behaviors that succeed only when guidance is present, rather than behaviors that transfer to the unguided setting. This provides evidence that overlap between guided and unguided state visitation, mediated by backtracking and revisiting intermediate steps, is an important component of \methodname{}'s efficacy.

\textbf{Qualitative evidence.}
We also compare model outputs produced by models trained with the default and modified instructions. As shown in Table~\ref{tab:prompt_qualitative_comparison}, under the default instruction, the unguided solution learned by \methodname{} often reflects concepts and intermediate steps that appear in guided traces (see Appendix~\ref{sec:qualitative_example} for the full problem, partial oracle solution, and representative guided/unguided rollouts), suggesting that the model stitches together reasoning learned under guidance. In contrast, with the modified instruction, the unguided solution no longer resembles the guided trace and instead follows a distinct solution path, exhibiting minimal reuse of concepts present in the guidance. This pattern is consistent with the intervention reducing overlap and thereby weakening the transfer mechanism.

\begin{table*}[ht]
  \centering
  \footnotesize
  \setlength{\tabcolsep}{6pt}
  \begin{tabularx}{\linewidth}{l cc}
    \toprule
    \textbf{Criterion} &
    \textbf{Training w/ Default Instruction} &
    \textbf{Training w/ Modified Instruction} \\
    \midrule
    Uses inequality structure
      & Yes
      & Yes \\
    Uses cyclic indexing meaningfully
      & Conceptual
      & Nominal only \\
    Uses $\lambda = \max S$ idea
      & Yes
      & Weak \\
    Follows partial response direction
      & Partial (extremal patterns)
      & No \\
    Uses geometric sequence $(a_i = x^{i-1})$
      & No
      & No \\
    Explores extremal constructions
      & Yes (patterns, ratios)
      & No \\
    Uses $(n \ge 4k)$ meaningfully
      & Partial
      & Mention only \\
    Depth of mathematical reasoning
      & Medium
      & Low \\
    True continuation of the partial response
      & Partial
      & No \\
    \bottomrule
  \end{tabularx}
  \vspace{-0.2cm}
  \caption{\footnotesize
  \textbf{Comparison of unguided solutions produced by models trained with the \methodname{} system instruction and the modified instruction on unguided and guided augmentations on hard problems}. Rollouts with the \methodname{} system instruction replicate several aspects of the guidance, indicating successful transfer. In contrast, rollouts from the modified system instruction show far lower resemblance to the guidance, suggesting that this instruction suppresses the stitching effect.}
  \vspace{-0.4cm}
  \label{tab:prompt_qualitative_comparison}
\end{table*}
\vspace{-0.2cm}
\section{Experimental Evaluation}
\label{sec:exps}
\vspace{-0.2cm}

The goal of our experiments is to evaluate the effectiveness of \methodname{} in solving hard problems during training and its impact on downstream performance. To this end, we address three main questions in this section: \textbf{(1)} Does \methodname{} improve the solvability of hard problems during training? \textbf{(2)} Does solving hard problems via \methodname{} improve performance on (potentially) out-of-distribution evaluation benchmarks? \textbf{(3)} How does \methodname{} compare to approaches that use oracle solutions as training targets, such as supervised fine-tuning on oracle solutions? We have already presented several diagnostic analyses in earlier sections, including how \methodname{} mitigates ray interference (Figure~\ref{fig:one_problem}) and the role played by the system instruction in enabling transfer (Section~\ref{sec:empirical_overlap}). We therefore focus mainly on performance results in this section.

\begin{wrapfigure}{r}{0.4\textwidth}
\vspace{-0.5cm}
\centering
\includegraphics[width=0.97\linewidth]{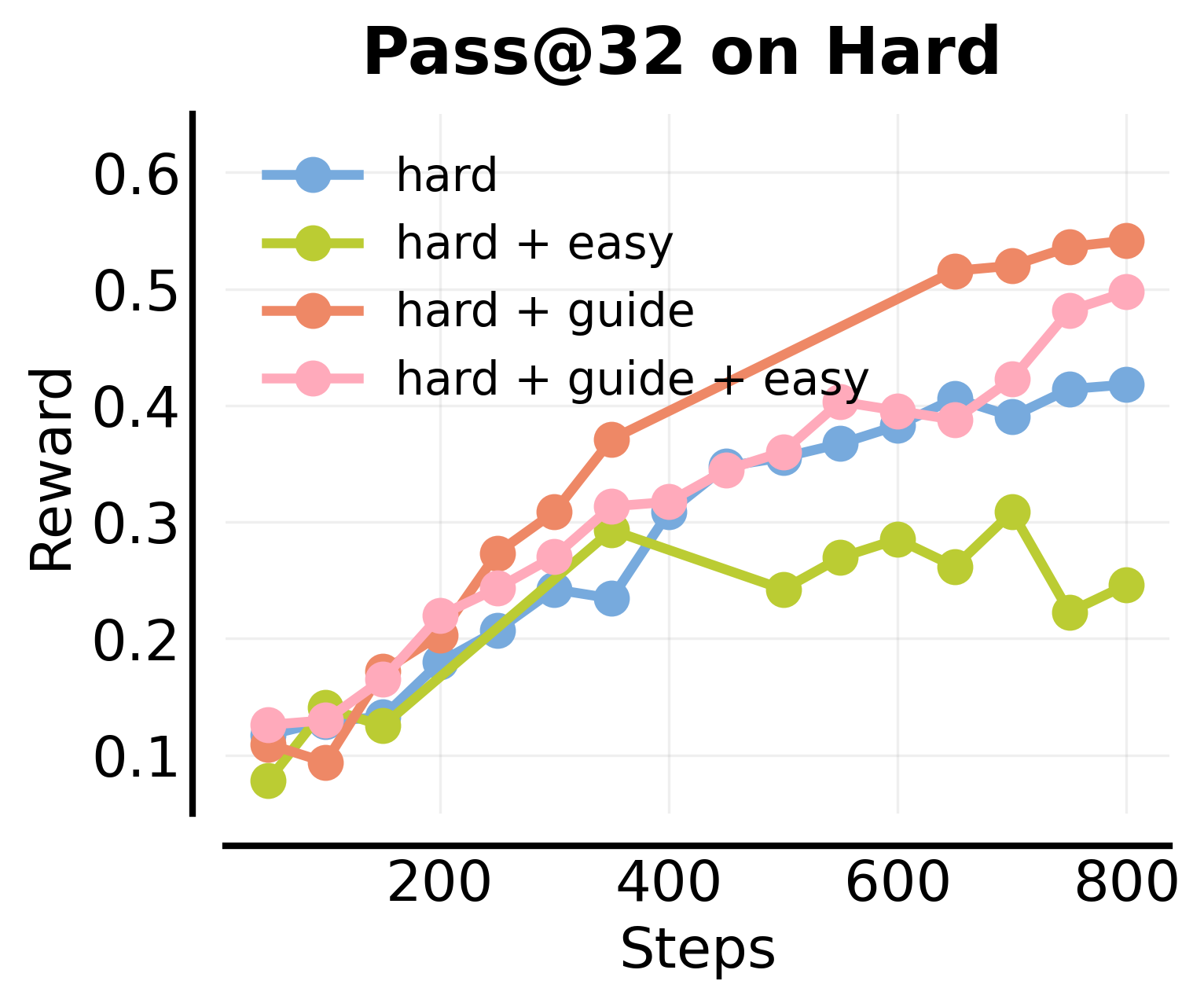}
\vspace{-0.3cm}
\caption{\footnotesize{\textbf{Pass@32 on the hard problem set evaluated with a 32k token budget.} Mixing in easy problems (green) causes a plateau in pass@32 over training, even though pass@32 continues to improve when training only on the hard set. This drop reflects ray interference caused by the easy data. In contrast, incorporating guidance in the form of a human-written prefix improves pass@32 consistently throughout training (red/pink), indicating that \methodname{} mitigates ray interference.}}
\vspace{-0.4cm}
\label{fig:pass@32_train}
\end{wrapfigure}

\textbf{Experimental setup.} We run all experiments using the \texttt{Qwen3-4B-Instruct-2507} base model. We train using GRPO with a maximum output length of 16384 tokens (recommended for this base model~\citep{yang2025qwen3}) and use a sampling temperature of 0.8. For most of our experiments, we use pipeline-rl~\citep{piche2025pipelinerl}, an asynchronous, streaming RL framework in our experiments, where we set the clip ratio of token-level importance weights to be $5.0$ on the higher end, and 0.0 on the lower end. We also implemented \methodname{} on verl~\citep{sheng2025hybridflow}, where we found similar preliminary results with 1 off-policy update step, and clip ratios of 0.2 and 0.28 on the lower and higher side respectively. Additional implementation details, hyperparameters, and details of our datasets are provided in Appendix~\ref{appendix: hyperparameters}. To construct the hard problem set, we select problems from ~\citep{yu2025dapo}, OmniMath (levels 5–8)~\citep{gao2024omnimathuniversalolympiadlevel}, and AceReason~\citep{chen2025acereason}. A problem is included only if the base model fails to produce any correct rollout under aggressive evaluation, using $k=128$ parallel samples and a 32k-token budget, ensuring that all selected problems lie in a near-zero–reward regime. Some examples are in Appendix~\ref{sec:dataset_example}.

\newcommand{\gainS}[1]{%
  \colorbox{red!15}{\scriptsize{#1}}%
}
\newcommand{\gainM}[1]{%
  \colorbox{red!30}{\scriptsize{#1}}%
}
\newcommand{\gainL}[1]{%
  \colorbox{red!45}{\scriptsize{#1}}%
}

\newcommand{\lossS}[1]{%
  \colorbox{yellow!15}{\scriptsize{#1}}%
}
\newcommand{\lossM}[1]{%
  \colorbox{yellow!30}{\scriptsize{#1}}%
}
\newcommand{\lossL}[1]{%
  \colorbox{yellow!45}{\scriptsize{#1}}%
}

\newcommand{\grp}[1]{\multicolumn{1}{l}{\emph{#1}}}
\newcommand{\ind}[1]{\hspace{1.2em}#1}

\textcolor{lightblue}{\textbf{Result 1: \methodname{} enables solving more hard problems.}} We first evaluate the efficacy of \methodname{} during training in Figure \ref{fig:pass@32_train}, where we evaluate the pass@32 performance on the training set under a much larger token budget of 32,768 tokens. Note that this evaluation configuration differs from training (which uses $8$ rollouts at 16,384 token length), and hence it stress tests if \methodname{} actually makes more progress on the training problems. Observe that \methodname{} (``hard + guide'') solves more problems from the hard set compared to any other configuration. While mixing in easy problems (``hard + easy'') results in a performance plateau on hard problems due to interference and this approach saturates at a lower pass@32 performance compared to training on hard problems alone (``hard''), no such performance plateau is observed for ``hard + guide'', which continues to improve as more steps of RL training are done. 

\textcolor{lightblue}{\textbf{Result 2: Training on broad problem mixtures with \methodname{}.}}
We further evaluate \methodname{} in a more practically relevant setting that mixes hard problems with varying amounts of easy problems, mimicking the broad training mixtures commonly used in practice. We report results in Table~\ref{tab:data_composition_results}. Concretely, we train on mixtures of ``hard + guide'' and the easy problem set, and compare against corresponding mixtures without guidance.
As shown in Figure~\ref{fig:pass@32_train} and Table~\ref{tab:data_composition_results}, mixing in an equal number of easy problems without guidance significantly degrades performance on the hard set (e.g., ``hard + easy'' vs.\ ``hard'') due to the interference issue discussed earlier. In contrast, \emph{introducing guidance via POPE substantially mitigates this problem}. For instance, even when easy problems are present, ``hard + guide + easy'' achieves a pass@1 of 14.3\% and pass@16 of 38.9\% on the hard set, which closely matches ``hard + guide'' alone (15.5\% pass@1 and 42.5\% pass@16), and does better than having no guidance.

This trend persists even when we scale up the amount of easy problems in the overall prompt set used for RL training. For instance, adding 1K easy problems without guidance severely harms hard-set performance (2.2\% pass@1), the corresponding mixture with guidance (``hard + guide + 1K easy'') recovers strong performance (14.0\% pass@1, 36.4\% pass@16), demonstrating that POPE enables robust learning on hard problems with diverse training mixtures.

\begin{table}[t]
\vspace{-0.3cm}
\centering
\small
\setlength{\tabcolsep}{5.5pt}
\renewcommand{\arraystretch}{1.05}

\resizebox{\linewidth}{!}{%
\begin{tabular}{@{}l c c c c c c@{}}
\toprule
\textbf{Approach} &
\multicolumn{2}{c}{\textbf{Hard problems}} &
\multicolumn{2}{c}{\textbf{AIME 2025}} &
\multicolumn{2}{c}{\textbf{HMMT 2025}} \\
\cmidrule(lr){2-3}\cmidrule(lr){4-5}\cmidrule(lr){6-7}
& pass@1 & pass@16 & pass@1 & pass@16 & pass@1 & pass@16 \\
\midrule

\grp{Base model} & & & & & & \\
\ind{Qwen3-4B-Instruct} 
& 0.57 & 7.42 & 48.13 & 77.29 & 29.06 & 52.99 \\
\midrule

\grp{Hard problems only} & & & & & & \\
\ind{+ hard}
& 13.55 & 32.89 & 49.58 & 81.43 & 31.04 & 63.79 \\
\ind{+ hard + guide (\methodname{})}
& 15.50 \gainS{+14\%} & 42.53 \gainM{+29\%}
& 53.12 \gainS{+7\%}  & 82.61 \gainS{+1\%}
& 37.81 \gainM{+22\%} & 67.49 \gainS{+6\%} \\
\ind{+ hard (Full-oracle SFT)} & 2.00 \lossL{$-$85\%} & 12.37 \lossL{$-$62\%} & 33.89 \lossM{$-$32\%} & 64.12 \lossM{$-$21\%} & 24.50 \lossM{$-$21\%} & 48.09 \lossM{$-$25\%}\\
\ind{+ hard (Prefix + rejection-sampled SFT)} & 5.14 \lossL{$-$62\%} & 24.50 \lossM{$-$26\%} & 38.12 \lossM{$-$23\%} & 77.62 \lossS{$-$5\%} & 30.08  \lossS{$-$3\%} & 50.91 \lossM{$-$20\%}\\
\midrule

\grp{Hard (256) + Easy (256)} & & & & & & \\
\ind{+ hard + easy}
& 8.22 & 23.81 & 57.19 & 82.50 & 37.19 & 62.81 \\
\ind{+ hard + guide + easy (\methodname{})}
& 14.32 \gainL{+74\%} & 38.93 \gainL{+63\%}
& 58.75 \gainS{+3\%}  & 83.87 \gainS{+2\%}
& 38.12 \gainS{+3\%}  & 67.15 \gainS{+7\%} \\
\midrule

\grp{Hard (256) + Easy (1K)} & & & & & & \\
\ind{+ hard + 1K easy problems}
& 2.24 & 25.12 & 61.88 & 83.79 & 37.03 & 60.07 \\
\ind{+ hard + guide + 1K easy problems (\methodname{})}
& 13.98 \gainL{+524\%} & 36.42 \gainM{+45\%}
& 62.01 \gainS{+0.2\%} & 84.31 \gainS{+0.6\%}
& 40.45 \gainM{+9\%}   & 70.38 \gainM{+17\%} \\
\bottomrule
\end{tabular}%
}
\vspace{-0.2cm}
\caption{\footnotesize \textbf{Pass@1 and pass@16 scores} on the hard set and standardized benchmarks (AIME2025 and HMMT2025), with relative gains \gainS{highlighted} in red and relative losses \lossS{highlighted} in yellow. Incorporating guidance via \methodname{} substantially improves performance on the hard problems while also improving performance on standardized benchmarks. The performance gains on hard problems are still preserved when a larger number of easy problems are mixed in. Mixing in easy problems improves performance on the easier AIME2025 benchmark, but training via \methodname{} on hard problems enables improvement on the harder HMMT2025 benchmark.}
\label{tab:data_composition_results}
\vspace{-0.3cm}
\end{table}

\textcolor{lightblue}{\textbf{Result 3: Performance on standardized benchmarks.}}
Table~\ref{tab:data_composition_results} reports performance on standardized benchmarks (AIME~2025 and HMMT~2025). 
Although guidance in \methodname{} is designed to target learning on hard training problems, it consistently improves downstream benchmark performance as well. 
In particular, ``{hard + guide + 1K easy}'' achieves the strongest overall results, obtaining the best pass@1 and pass@16 on both AIME~2025 and HMMT~2025 under a 32{,}768-token evaluation budget. These gains arise because the easy dataset overlaps in difficulty with standardized benchmarks, while \methodname{} enables learning on harder problems despite a substantial ratio of easy problems in the mixture.
Most notably, the improvement from adding guided hard problems is larger on HMMT~2025, a benchmark typically considered harder than AIME~2025 (e.g., ``{hard + guide + 1K easy}'' vs.\ ``{hard + 1K easy}''; $+3.4\%$ in pass@1 and $+10.0\%$ on pass@16). This comparison highlights the benefit of maintaining effective learning on hard problems, even though there are sufficiently many easy problems in the training mixture. Overall, these results demonstrate that \methodname{} not only improves optimization on difficult training instances, but also scales robustly to large, heterogeneous data mixtures commonly used in practice.

\textcolor{lightblue}{\textbf{Result 4: Comparison with methods that use oracle solutions as training targets.}} Finally, we compare the performance and optimization behavior of \methodname{} with prior approaches that use oracle solutions directly as training targets. Specifically, we compare against two methods that apply supervised fine-tuning (SFT) on privileged information followed by standard RL. We also attempted to compare to LUFFY~\citep{yan2025learning}, which incorporates the oracle solution directly as a rollout during RL but were unable to make it train stably on our hard problems with human reference solutions; hence we skip this comparison for now. The two SFT baselines are: \textbf{(a)} SFT directly on the oracle solution (``Full-oracle SFT''), and \textbf{(b)} SFT on the prefix of the oracle solution followed by a successful on-policy completion obtained via rejection sampling from the base model (called ``Prefix + rejection-sampled SFT''). Note that the oracle prefixes used by these baselines are identical to those employed by \methodname{}.

As shown in Table~\ref{tab:data_composition_results}, SFT on full oracle solutions severely degrades performance across all evaluations. On the hard problem set, this approach reduces pass@1 from $13.6\%$ (``+ hard'') to $2.0\%$ and pass@16 from $32.9\%$ to $12.4\%$. This performance degeneration also manifests on standardized benchmarks, with AIME~2025 pass@1 decreasing from $49.6\%$ to $33.9\%$ and HMMT~2025 pass@16 falling from $63.8\%$ to $48.1\%$.  This is perhaps expected since oracle solutions exhibit fundamentally different reasoning styles and cloning such off-policy data disrupt the model's own reasoning capabilities~\citep{yang2026intselfproposedinterventionsenable}. 

The rejection-sampled SFT variant avoids catastrophic collapse but still underperforms RL training on hard problems. Specifically, it achieves only $5.1\%$ pass@1 and $24.5\%$ pass@16 on the hard set, substantially below both ``+ hard'' ($13.6\%$ / $32.9\%$) and ``+ hard + guide'' ($15.5\%$ / $42.5\%$). While its performance on easier benchmarks such as AIME~2025 remains close to the base model, it even underperforms na\"ive RL training on hard problems starting from the base model (``hard''). In Appendix~\ref{app:sft_plus_rl}, we further show that applying RL on top of the SFT warm start does not improve exploration, yielding virtually no gains in hard-problem solvability compared to standard RL (``hard'').

\begin{AIbox}{Takeaways: \methodname{} enables robust learning on hard problems}
\begin{itemize}[itemsep=0pt]
 \setlength{\leftskip}{-15pt}
    \item \methodname{} consistently improves the solvability of hard problems, avoiding ray interference.
    \item \methodname{} preserves strong performance on hard problems even when training with easy problems.
    \item Training on hard problems via \methodname{} also improves performance on standardized benchmarks.
\end{itemize}
\end{AIbox}

\vspace{-0.2cm}
\section{Related Work}
\label{sec:related}
\vspace{-0.2cm}

We tackle exploration on hard problems in regimes where na\"ively scaling on-policy RL compute yields little progress. At its core, this is an \emph{exploration} challenge, since algorithmic interventions are required to discover high-reward trajectories. We therefore briefly discuss related approaches for improving exploration, including methods that add explicit exploration bonuses and methods that learn from off-policy traces that cover high-reward regions. We also discuss connections with ideas from RL theory.

\textbf{Exploration methods in RL.} Recent work has shown that reinforcement learning can substantially improve LLM reasoning by reinforcing long-horizon behaviors such as self-correction and reflection~\citep{liu2025prorlprolongedreinforcementlearning,deepscaler2025,qu2024recursive,gandhi2025cognitivebehaviorsenableselfimproving}. However, multiple studies observe that on-policy RL tends to over-optimize already-solvable problems, leaving harder problems unsolved~\citep{yue2025doesreinforcementlearningreally,zhao2025echochamberrlposttraining}. At the population level, this often manifests as declining pass@$k$ despite increasing training reward. As we show in this work, this behavior can be explained by ray interference~\citep{schaul2019ray}, which biases optimization toward states where reward is already attainable, creating a structural barrier to learning on hard problems. To address ray interference, our approach \methodname{} makes it possible to make more ``uniform'' updates on all problems by incorporating guidance derived from a human-written solution (available in most datasets). 

Several prior approaches attempt to mitigate over-sharpening using exploration bonuses~\citep{gao2025navigateunknownenhancingllm,wang2025reinforcementlearningreasoninglarge,hamid2025polychromic,song2025outcomebasedexplorationllmreasoning}, objectives that directly optimize pass@$k$~\citep{chow2024inference,balashankar2025infaligninferenceawarelanguagemodel}, or curricula and prompt mixtures that rely on transfer from easier problems~\citep{setlur2025e3learningexploreenables,sun2025rl,liu2025prorlprolongedreinforcementlearning,hu2025brorl}. However, these methods fundamentally depend on sampling at least one correct rollout. When pass@1 is near zero, token-level exploration provides no useful signal, pass@$k$ optimization reduces to a monotonic transformation of pass@1, and transfer from easier problems fails due to interference. Our experiments in Section~\ref{sec:rich_gets_richer} showcase failure modes of a representative subset of these approaches in addressing the challenge of learning on hard problems. 

Our approach is conceptually related to classical RL results showing that access to intermediate states or resets can significantly reduce the sample complexity of exploration~\citep{jaksch2010near,azar2017minimax,kakade2002approximately,agarwal2021theory}, as well as modern methods such as Go-Explore~\citep{ecoffet2019go,ecoffet2020return} that revisit previously discovered states. Unlike these methods, \methodname{} does not perform hard resets or rely on explicit state visitation. Instead, we leverage the instruction-following capabilities of LLMs to steer on-policy rollouts into analogous internal states that enable learning signal, and crucially, allow behaviors learned under guidance to transfer back to unguided problems. To our knowledge, prior work does not systematically study this guided-to-unguided transfer mechanism or explain why guided training can improve performance when guidance is absent at test time. 

Although we do not establish formal theoretical guarantees for LLMs in this work, we note that \methodname{} may violate several standard assumptions underlying these results, including realizability of oracle demonstrations, uniform coverage via random prefix sampling at every learning step, and the absence of update interference, which plays a central role in our analysis. This suggests that new abstractions may be required to theoretically study exploration in LLMs, an important direction for future work.

\textbf{Training LLMs on off-policy traces.}
Motivated by the limitations of on-policy RL, several works propose updating LLM policies using human- or oracle-provided reasoning traces~\citep{lightman2023lets,corrado2024guideddataaugmentationoffline}. While effective in some settings, methods that rely on supervision from a teacher model are inherently bounded by the teacher’s capacity~\citep{agarwal2024onpolicydistillationlanguagemodels}. Moreover, stable learning from off-policy traces often requires additional mechanisms such as reward shaping~\citep{yan2025learningreasonoffpolicyguidance}, entropy control~\citep{wang2025beyond}, and careful hyperparameter tuning~\citep{zhang2025onpolicyrlmeetsoffpolicy}.

A more fundamental limitation is that suitable off-policy reasoning traces are not readily available for many hard problems. Although human-written solutions exist for most training prompts and can often be rephrased into more effective formats, producing long chains of thought that align with how models actually reason remains challenging~\citep{zelikman2022star}. This mismatch between off-policy traces and the model’s native reasoning behavior can lead to unstable learning dynamics, including entropy explosion or entropy collapse depending on the SFT configuration, as discussed conceptually in Section~\ref{sec:pope}. These limitations motivate approaches that avoid using off-policy traces as direct training targets.

\textbf{Most related prior works.} The most closely related prior and concurrent works that address learning on hard problems leverage human or oracle data to extract subgoals, plans, or abstractions, which are then used to inform rollout generation in on-policy RL~\citep{hong2025planning,qu2025learning,li2025questa,chen2025nudging}. Our work shares a similar high-level philosophy, but shows that simply conditioning on prefixes of past solutions is sufficient to enable learning on hard problems. Prior work that also directly utilizes partial solutions~\citep{amani2025rlreasoningadaptivelyrevealing,zhang2025bread} primarily studies non-reasoning models that produce short responses and focuses on problems that are not too hard. In particular, with non-reasoning models, \citet{amani2025rlreasoningadaptivelyrevealing} requires adaptively tuning the length of the partial solution for on-policy generation, whereas we find that \methodname{} does not require such curricula, since backtracking and recovery behaviors naturally provide coverage over states close to initialization.

Moreover, to the best of our knowledge, no prior work systematically studies why unguided training is difficult on hard problems, identifies the limitations of existing approaches, and establishes the role of guided training in enabling transfer to settings where guidance is absent. It is in fact unclear many times from prior work, why a guided approach is needed in the first place and simple adjustments to learning configurations of existing algorithms are insufficient. In contrast, we identify the ray interference problem, show that it cannot be solved by several token-level exploration or pass@k optimization approaches, and develop a mental model under which training on augmented problems enables transfer to unguided problems (Section~\ref{sec:empirical_overlap}). We validate all these insights through targeted empirical interventions.

\vspace{-0.3cm}
\section{Discussion and Perspectives on Future Work}
\vspace{-0.25cm}

In this paper, we study a fundamental limitation of on-policy RL for LLMs: the inability to learn from hard problems when no correct rollouts are sampled. We show that standard remedies for exploration, including entropy bonuses, optimistic updates, pass@$k$ optimization, and curricula over easy problems, fail to address this challenge due to sharpening and ray interference. To overcome this limitation, we introduce \methodfullname{} (\methodname{}), a framework that leverages privileged information in the form of partial oracle solutions to guide on-policy exploration without using these solutions as training targets. By conditioning rollouts on solution prefixes along with a system instruction to build on this prefix, and training on a mixture of guided and unguided problems, \methodname{} enables the model to obtain learning signal on hard problems and acquire reasoning behaviors that transfer back to unguided settings. We provide empirical results showing that this transfer is enabled by a synergy between instruction-following and reasoning behaviors, and demonstrate that \methodname{} substantially expands the set of solvable hard problems where existing RL approaches fail.

There are several directions for future work. \textbf{First}, formalizing the mechanism by which \methodname{} improves exploration on hard problems is an important open question. Our experiments suggest that \methodname{} improves performance by leveraging the instruction-following capabilities of the underlying LLM to follow and build upon oracle solutions. How can this notion be quantified theoretically? From a practical perspective, how can these instruction-following capabilities be further amplified and systematically leveraged to improve reasoning? \textbf{Second}, there exists a class of even harder problems for which models fundamentally lack the knowledge required to solve the task. In such cases, conditioning on an oracle solution and relying on instruction following alone may be insufficient to improve performance, and deriving explicit training targets from the oracle may be necessary. How should such training targets be constructed? How can we mitigate challenges associated with memorization and pathological optimization in this regime? We believe that methods from off-policy RL, for example, training explicit value functions~\citep{setlur2024rewarding,setlur2025opt}) or implicitly modeling them via interventions~\citep{yang2026intselfproposedinterventionsenable} likely provide a natural starting point for answering this question. \textbf{Third}, our work highlights the role of ray interference in inhibiting learning on heterogeneous prompt mixtures. Ray interference is not unique to our prompt sets, is a more general phenomenon that is likely present a bigger prompt sets as well. What factors determine the severity of this interference? How does it depend on the model’s pre-training or mid-training procedures? Can we predict when ray interference will arise before running RL training? Addressing these questions will lead to more robust and predictable RL recipes that continue to make progress without prematurely plateauing on heterogeneous dataset mixtures.

\vspace{-0.2cm}
\section*{Acknowledgements}
\vspace{-0.2cm}

We thank Matthew Yang, Zheyuan Hu, Max Sobol Mark, Anikait Singh, Rafael Rafailov, Apurva Gandhi, and others in the CMU AIRe lab for discussions and feedback. This work is supported by the Office of Naval Research under N0014-24-2206 and a Schmidt Sciences AI2050 Early Career Fellowship. We thank the Orchard cluster at the CMU FLAME center for most of the GPU resources that powered this work, and DeltaAI for providing compute support for some of the critical experiments in this paper. We thank TPU research cloud (TRC) for their generous support. YQ gratefully acknowledges the support of the Amazon AI PhD Fellowship; AS gratefully acknowledges the support of JP Morgan AI Fellowship.

\bibliography{main}

\newpage
\appendix
\onecolumn

\newpage

\part*{Appendices}

\vspace{-0.2cm}
\section{Why Does Entropy Increase with a Higher Clip Ratio?}
\label{app:entropy_clip_ratio}
\vspace{-0.2cm}

We now briefly attempt to understand the mechanism behind our finding that increasing the clip ratio, as in DAPO~\citep{yu2025dapo}, can lead to higher next-token entropy even without an explicit entropy regularizer. Off-policy negative samples generally push the model toward higher token entropy on average, as shown theoretically and empirically by \citet{setlur2025e3learningexploreenables}. Increasing the positive clip ratio amplifies this effect on hard problems by allowing more optimistic policy updates on low-likelihood positive traces.
By definition, the base model is unlikely to sample a successful trace on a hard problem, so positive trajectories are rare and assigned very low probability under the current policy. A larger clip ratio permits updates that attempt to increase the likelihood of tokens appearing in rare traces; however, with only one or a few gradient steps, the model cannot fully reallocate probability mass onto them. As a result, some tokens in rare positive traces receive a disproportionate increase in probability mass. In reasoning models, these are often tokens that signal a shift in the reasoning trajectory (e.g., ``Wait'', ``maybe''), which are known to induce high-entropy next-token distributions~\citep{wang2025beyond}. Instead, it reduces confidence in previously high-probability tokens without successfully fitting the positive trace, resulting in a flatter next-token distribution and increased entropy. Repeating this process across training steps leads to an entropy explosion. Had the update been able to fully concentrate mass on the positive trace, or been suppressed entirely, this entropy amplification would not occur.

\vspace{-0.2cm}
\section{Details of Pass@$k$ Policy Optimization}
\label{app:pass_at_k_more}
\vspace{-0.2cm}

In Section~\ref{sec:ray_int}, we experimented with pass@$k$ optimization to see if it can solve the ray interference problem by not over-optimizing pass@1. Recall that ray interference is a direct consequence of the competition between optimizing reward on problems where rewards can already be attained and optimizing reward on new problems. Naturally, one might except that if only optimize pass@$k$ for a higher value of $k$ (e.g., $k=8$), then we may no longer run into the issue of over-optimizing pass@1 on some problems at the cost of performance and increase the hardness of sampling a correct rollout on the others. Here, we detail the objective we use for pass@$k$ optimization from prior work~\citep{walder2025pass}.

\textbf{Pass@$k$ estimator.}
Given a prompt $\bx$, we sample $n \ge k$ i.i.d.\ rollouts $\by_1,\ldots,\by_n \sim \pi_\theta(\cdot\mid \bx)$ and evaluate their correctness $f_i \triangleq R(\bx,\by_i) \in \{0,1\}$. Let $c \;\triangleq\; \sum_{i=1}^n f_i $
denote the number of correct rollouts in the batch. An unbiased estimator of pass@$k$ objective is given by:
\begin{align}
\rho(n,c,k)
\;\triangleq\;
1 - \frac{\binom{n-c}{k}}{\binom{n}{k}},
\label{eq:passk_estimator}
\end{align}
which estimates the probability that at least one of $k$ uniformly sampled rollouts (without replacement) is correct. Intuitively, $\rho(n,c,k)$ increases monotonically with the number of observed successes $c$, and reduces to pass@1 when $k=1$.

In our setting, this estimator is applied at the level of individual prompts, and the overall training objective is to maximize the expected pass@$k$ score across the training distribution:
\begin{align}
\mathcal{J}_{k}(\theta)
\;\triangleq\;
\mathbb{E}_{\bx \sim \rho}\!\left[
\mathbb{E}_{\by_1,\ldots,\by_n \sim \pi_\theta(\cdot\mid\bx)}
\big[
\rho(n,c(\bx),k)
\big]
\right],
\label{eq:passk_objective}
\end{align}
where $\rho$ denotes the empirical distribution over training prompts and $c(\bx)$ is the number of correct rollouts for prompt $\bx$.

\textbf{Unbiased pass@$k$ gradient estimator.}
With this definition, we now present the policy gradient term that we use from \citet{walder2025pass}. Given a prompt $\bx$, sample $n$ i.i.d.\ rollouts $\by_1,\ldots,\by_n \sim \pi_\theta(\cdot\mid \bx)$ with correctness labels
$f_i \triangleq R(\bx,\by_i)\in\{0,1\}$ and let $c \triangleq \sum_{i=1}^n f_i$ be the number of correct samples.
An unbiased estimator of the gradient of the (per-prompt) pass@$k$ objective can be written as a weighted policy-gradient update:
\begin{align}
\widehat{\nabla_\theta}
\;=\;
\sum_{i=1}^n r_i \,\nabla_\theta \log \pi_\theta(\by_i \mid \bx),
\label{eq:passk_grad_est}
\end{align}
where the weights are
\begin{align}
r_i
=
\begin{cases}
\frac{k}{n}, & \text{if } f_i = 1,\\[4pt]
\frac{k}{n}\,\rho(n-1,c,k-1), & \text{if } f_i = 0,
\end{cases}
\label{eq:passk_weights_ri}
\end{align}
and $\rho(\cdot)$ is the unbiased pass@$k$ estimator from~\eqref{eq:passk_estimator}:
\begin{align}
\rho(n,c,k)=1-\frac{\binom{n-c}{k}}{\binom{n}{k}}.
\end{align}
We can now treat each of the values in Equation~\ref{eq:passk_weights_ri} as ``reward'' and run standard RL to optimize it. We chose to use this instantiation of the pass@$k$ policy optimization objective over the variant of \citet{chow2024inference} because this version is simpler in terms of implementation.

\vspace{-0.2cm}
\section{Why does SFT + RL Not Improve Solvability on Hard Problems?}
\label{app:sft_plus_rl}
\vspace{-0.2cm}

\begin{wrapfigure}{r}{0.6\textwidth}
\vspace{-0.5cm}
\centering
~~~~\includegraphics[width=0.97\linewidth]{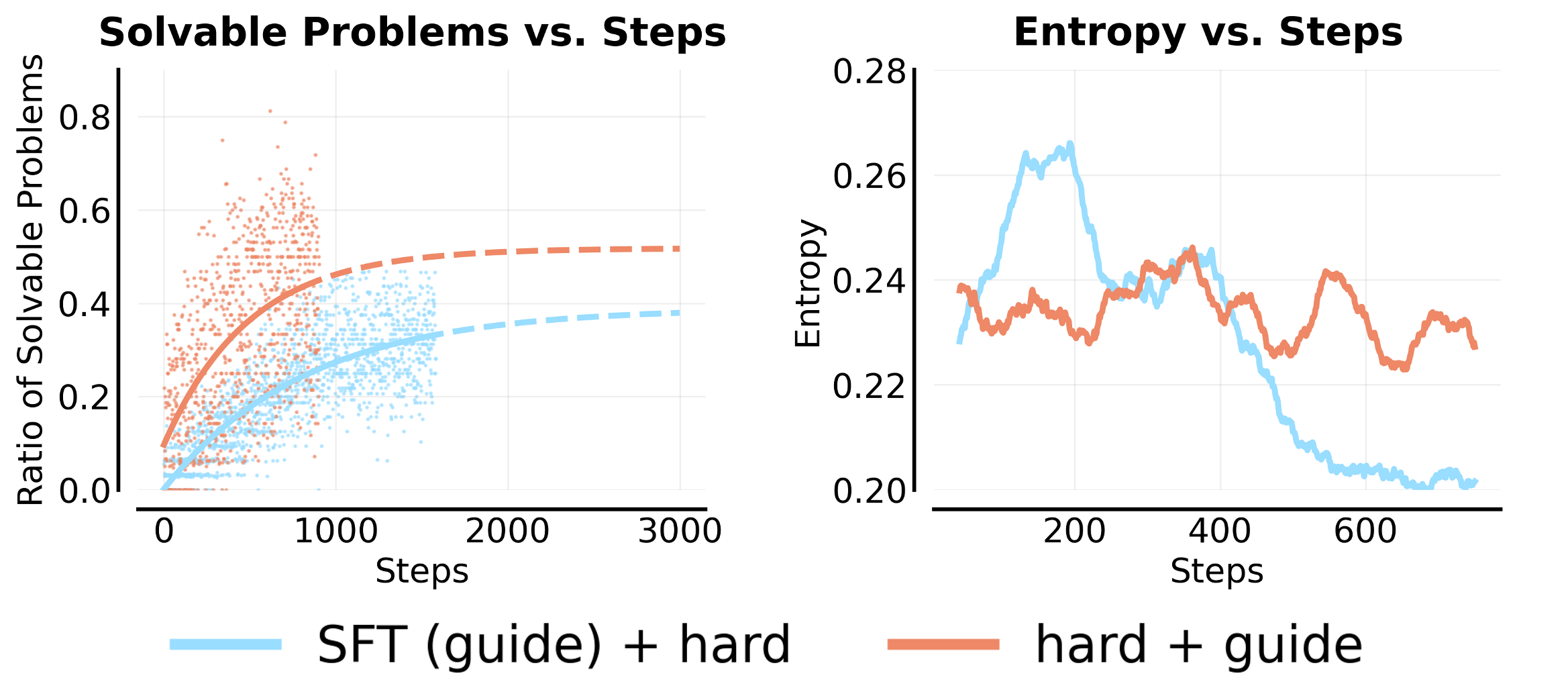}
\vspace{-0.6cm}
\caption{\footnotesize{
\textbf{Effect of SFT warmstarts on solvability and entropy.}
\textbf{Left:} Fraction of solvable hard problems during training.
Warm-starting RL from an SFT model trained on synthetically generated, rejection-sampled traces results in consistently worse solvability than our approach. \textbf{Right:} SFT warmstarts induce a persistent entropy collapse, leading to reduced exploration and suboptimal on-policy learning.
}}
\vspace{-0.5cm}
\label{fig:sft_rl}
\end{wrapfigure} 

Although warmstarting with SFT is often effective when high-quality expert traces are available, it fundamentally alters the reasoning behaviors of the base model, leading to poor performance in Table~\ref{tab:data_composition_results}. Even when SFT is restricted to a short prefix of the oracle solution (as used for \methodname{}), followed by a correct on-policy reasoning trace obtained via rejection sampling, SFT concentrates probability mass onto a narrow set of token-level distributions. As shown in Figure~\ref{fig:sft_rl}, initializing RL from such an SFT-trained checkpoint leads to a collapse in token entropy and substantially worse solvability compared to our approach (\methodname{}; ``hard + guide'').

More broadly, low-entropy initialization is especially harmful in sparse-reward regimes. On hard problems, successful trajectories are rare and lie in the tail of the policy’s distribution; once entropy collapses, on-policy sampling rarely explores alternative reasoning paths, and policy-gradient updates become dominated by near-duplicate prefixes. As a result, the policy becomes trapped in a locally consistent but globally suboptimal mode, preventing progress on previously unsolved problems.

\vspace{-0.2cm}
\section{Extended Discussion of the Overlap Hypothesis}
\label{app:extended_overlap}
\vspace{-0.2cm}

The core intuition is that \methodname{} converts a sparse-reward exploration problem into a two-stage problem with a much easier first stage. In the MDP picture (Figure~\ref{fig:mental_model}), the bottleneck is not improving behavior \emph{within} the reward-bearing region, but rather \emph{reaching} any state from which reward is attainable. The guidance (or prefix) functions as a roll-in distribution that reliably lands the learner in $\mathcal{S}_{\text{good}}$, so that even early in training the algorithm can observe non-zero reward and fit effective continuations. Once the continuation policy is learned, the role of guidance is largely complete: it is no longer needed to succeed \emph{from} $\mathcal{S}_{\text{good}}$, but it has created a bank of successful trajectories that certify which parts of the state space admit reward and what actions to take there. This “certification” is crucial because membership in $\mathcal{S}_{\text{good}}$ is only revealed by downstream success; without guided roll-ins, standard on-policy RL must simultaneously discover $\mathcal{S}_{\text{good}}$ and learn to exploit it, which is exponentially harder when reward is rare.

\textbf{Why can this transfer work for LLM reasoning?} In autoregressive LLMs, a state can be viewed as the generated prefix, but for long chain-of-thought reasoning the more relevant notion is the model’s internal representation induced by that prefix. Due to self-correction and backtracking, many distinct token sequences can correspond to similar latent ``problem-solving states''. Guidance therefore helps not only by increasing the likelihood of a successful continuation, but by steering the model into internal states that are both \emph{reachable} and \emph{stable} under subsequent on-policy sampling.
When guided rollouts exhibit behaviors such as self-verification, restarting, or revisiting earlier steps, they induce overlap between guided states and a neighborhood of states that the unguided policy can plausibly reach on its own. \textbf{Under \emph{function approximation}}, this overlap allows learning signal from guided successes to generalize to unguided prefixes, effectively reducing the remaining exploration problem to reaching any nearby state rather than reproducing the full guidance string. This perspective also clarifies why transfer can occur even with a fixed segment of guidance: the structure of reasoning traces induces many revisitations and near-collisions in latent state space, so a single guided roll-in can populate a broad set of useful states from which the learned continuation policy can succeed.

\vspace{-0.3cm}
\section{Training Hyperparameters}
\vspace{-0.2cm}
\label{appendix: hyperparameters}

This section summarizes the training hyperparameters and system configuration used in this paper. We additionally describe key components of the \texttt{Pipeline-RL} framework that orchestrates distributed rollout, preprocessing, and optimization.
\begin{table*}[ht]
\centering

\begin{minipage}[t]{0.48\linewidth}
\raggedright
\centering
\vspace{0pt} 
\begin{tabularx}{\linewidth}{l|c}
  \toprule
\multicolumn{1}{c}{\textbf{Hyperparameter}} \vline &
\multicolumn{1}{c}{\textbf{Values}} \\
\midrule
learning rate & $1.0 \times 10^{-5}$ \\
num. train epochs & 3 \\
global batch size & 128 \\
gradient checkpointing & True \\
max sequence length & 16384 \\
precision & bf16 \\
num. GPUs & 8 \\
warmup ratio & 0.1 \\
\bottomrule
\end{tabularx}
\caption{Hyperparameters used for SFT.}
\label{tab:finetune_hyper}
\end{minipage}
\hfill
\begin{minipage}[t]{0.48\linewidth}
\raggedright
\centering
\vspace{0pt} 
\begin{tabularx}{\linewidth}{l|c}
  \toprule
\multicolumn{1}{c}{\textbf{Hyperparameter}} \vline &
\multicolumn{1}{c}{\textbf{Values}} \\
\midrule
max prompt length & 2048 \\
max response length & 16384 \\
sampling temperature & 0.8 \\
clip ratio (low / high) & 0 / 5.0 \\
train batch size & 32 \\
learning rate & $1.0 \times 10^{-6}$ \\
total training steps & 1000 \\
num. GPUs & 8 \\
rollouts per prompt & 8 \\
ring buffer size & 128 \\
\bottomrule
\end{tabularx}
\caption{Hyperparameters used for RL training with GRPO under the Pipeline-RL framework.}
\label{tab:finetune_hyper-2}
\end{minipage}

\end{table*}

\subsection{Hyperparameters for SFT}
\label{sec:hyper-open}

For supervised fine-tuning (SFT), we use the \href{https://github.com/huggingface/trl}{TRL} codebase. All models are initialized from \href{https://huggingface.co/Qwen/Qwen3-4B-Instruct-2507}{Qwen3-4B-Instruct}. Training is performed with full-parameter fine-tuning using bfloat16 precision and gradient checkpointing.

\subsection{Hyperparameters for RL}
\label{sec:hyper-open-2}

For reinforcement learning, we adopt the \href{https://github.com/ServiceNow/PipelineRL}{\texttt{Pipeline-RL}} framework with GRPO as the underlying optimization algorithm. At a high level, the training pipeline consists of (i) actor workers that generate rollouts, (ii) preprocessing workers that chunk, filter, and buffer samples, and (iii) learner workers that perform policy optimization. Actors generate up to 8 rollouts per prompt, and samples are stored in a ring buffer with capacity 128 to replace stale data when training lags behind generation. Same as SFT, we use \href{https://huggingface.co/Qwen/Qwen3-4B-Instruct-2507}{Qwen3-4B-Instruct} as the base policy model.

\vspace{-0.2cm}
\section{Qualitative Example}
\label{sec:qualitative_example}

In this section, we present a representative example used in the analysis of Section~\ref{sec:empirical_overlap}. We first show the original problem, and the partial human-written solution provided as guidance during training.

\begin{promptbox}{Problem}
\small
Given positive integers $n, k$ such that $n\ge 4k$, find the minimal value $\lambda=\lambda(n,k)$ such that for any positive reals $a_1,a_2,\ldots,a_n$, we have \[ \sum\limits_{i=1}^{n} {\frac{{a}_{i}}{\sqrt{{a}_{i}^{2}+{a}_{{i}+{1}}^{2}+{\cdots}{{+}}{a}_{{i}{+}{k}}^{2}}}} \le \lambda\] Where $a_{n+i}=a_i,i=1,2,\ldots,k$
\end{promptbox}

\begin{promptbox}{Partial Oracle Solution (Used as Guidance during Training)}
\small
Let $S$ be the sum in question: \[ S = \sum\limits_{i=1}^{n} {\frac{{a}_{i}}{\sqrt{{a}_{i}^{2}+{a}_{{i}+{1}}^{2}+{\cdots}{{+}}{a}_{{i}{+}{k}}^{2}}}} \] Let $D_i = {a}_{i}^{2}+{a}_{{i}+{1}}^{2}+{\cdots}{{+}}{a}_{{i}{+}{k}}^{2}$. So the sum is $S = \sum_{i=1}^n \frac{a_i}{\sqrt{D_i}}$. The value $\lambda(n,k)$ is the maximum possible value of $S$. 

Step 1: Find a lower bound for $\lambda$. We will construct a sequence $(a_i)$ for which the sum $S$ approaches a certain value. This value will be a lower bound for $\lambda$. Let $a_i = x^{i-1}$ for $i=1, 2, \ldots, n$ for some small positive real number $x$. The indices are cyclic, so $a_{n+j} = a_j = x^{j-1}$.
\end{promptbox}

We next show representative unguided solutions generated by models trained with different system instructions. During evaluation, no guidance or partial oracle solution is provided to the model.

\begin{promptbox}{Unguided Solution (Model Trained with Default Instruction)}
\small
We are given positive integers $ n, k $ with the condition $ n \geq 4k $, and we are to find the minimal value $ \lambda = \lambda(n,k) $ such that for any positive real numbers $ a_1, a_2, \ldots, a_n $ (with indices modulo $ n $, so $ a_{n+i} = a_i $), the following inequality holds:
\[
\sum_{i=1}^n \frac{a_i}{\sqrt{a_i^2 + a_{i+1}^2 + \cdots + a_{i+k}^2}} \leq \lambda.
\]

Our goal is to find the minimal such $ \lambda $ that works for all such sequences, under the constraint $ n \geq 4k $.

\ldots

Each term in the sum is of the form
\[
\frac{a_i}{\sqrt{a_i^2 + a_{i+1}^2 + \cdots + a_{i+k}^2}}
= \frac{1}{\sqrt{1 + \sum_{j=1}^k \left(\frac{a_{i+j}}{a_i}\right)^2}}.
\]

\ldots

Suppose we try to make $ a_{i+1}, a_{i+2}, \ldots, a_{i+k} $ very small compared to $ a_i $. Then
\[
\frac{a_i}{\sqrt{a_i^2 + \sum_{j=1}^k a_{i+j}^2}} \approx 1.
\]

But due to the cyclic structure, each $ a_j $ appears in multiple denominators, so increasing one term necessarily affects others.

\ldots

So there is a trade-off between making individual terms large and maintaining consistency across the cycle. This suggests that extremal or periodic constructions may be relevant in maximizing the sum.

\ldots
\end{promptbox}

\begin{promptbox}{Unguided Solution (Model Trained with Modified Instruction)}
\small
We are given positive integers $ n, k $ with $ n \geq 4k $, and we are to find the minimal value $ \lambda = \lambda(n,k) $ such that for any positive real numbers $ a_1, a_2, \ldots, a_n $ (with indices modulo $ n $), the following inequality holds:
\[
\sum_{i=1}^{n} \frac{a_i}{\sqrt{a_i^2 + a_{i+1}^2 + \cdots + a_{i+k}^2}} \leq \lambda.
\]

\ldots

Suppose all $ a_i $ are equal. Let $ a_i = 1 $ for all $ i $. Then each denominator is $ \sqrt{k+1} $, and each term equals $ 1/\sqrt{k+1} $. The total sum is therefore
\[
\frac{n}{\sqrt{k+1}}.
\]

\ldots

Each term is at most $ 1 $, since the denominator includes $ a_i^2 $. Thus the entire sum is bounded above by $ n $.

\ldots

We can attempt to apply Cauchy--Schwarz or symmetry arguments to bound the sum, but these approaches do not immediately yield a tighter characterization of the extremal behavior.

\ldots

\end{promptbox}

\section{More Examples}
\label{sec:dataset_example}
Here we provide an example from the Omni-MATH dataset with a human solution.
\begin{promptbox}{Question from Omni-MATH}
\small
Let $k\ge2$ be an integer. Find the smallest integer $n \ge k+1$ with the property that there exists a set of $n$ distinct real numbers such that each of its elements can be written as a sum of $k$ other distinct elements of the set.
\end{promptbox}

\begin{promptbox}{Human Solution}
\small
Let \( k \geq 2 \) be an integer. We need to find the smallest integer \( n \geq k+1 \) such that there exists a set \( S \) of \( n \) distinct real numbers, where each element of \( S \) can be expressed as a sum of \( k \) other distinct elements of \( S \).

To solve this problem, we consider the construction of such a set \( S \).

1. **Understanding the Problem:**

- For each element \( s \in S \), we need \( k \) distinct elements from \( S \setminus \{s\} \) that sum up to \( s \).

2. **Minimum Size Construction:**

- We start by proving that with \( n = k + 4 \), such a set can indeed be constructed.

- Consider a construction where:

- Choose \( k + 1 \) elements as the base set: \(\{ a_1, a_2, \ldots, a_{k+1} \} \).

- Introduce an additional four elements: \(\{ b_1, b_2, b_3, b_4 \} \).

- We construct our set \( S \) as:
\[
S = \{ a_1, a_2, \ldots, a_{k+1}, b_1, b_2, b_3, b_4 \}
\]

3. **Illustrating the Construction:**

- Arrange the elements such that:

- Each \( a_i \) is expressed as the sum of any \( k \) of the other \( a_j \)'s and some \( b \)'s if necessary.

- Each \( b_i \) can be expressed using a combination of \( a \)'s and other \( b \)'s.

4. **Verification:**

- By choosing specific numbers for each \( b_i \), we ensure that each number in the constructed set can indeed be expressed as a sum of \( k \) distinct others.

- For example, by choosing values and testing that the sum condition holds, we verify that each possibility works, fulfilling the problem's conditions.

5. **Conclusion:**

- Testing smaller \( n \) for valid configurations will fail due to insufficient numbers to formulate each possible sum using \( k \) distinct numbers.

- Therefore, the smallest \( n \) for which such a configuration is possible indeed turns out to be \( n = k + 4 \).

Thus, the smallest integer \( n \) such that a set \( S \) with the given conditions can be constructed is: $\boxed{k + 4}$
\end{promptbox}

Here we provide an example from the DAPO dataset with a solution generated by gemini-2.5-pro.

\begin{promptbox}{Question from DAPO}
\small
Find the smallest natural number $n$ such that the expression $n^2 - n + 11$ has exactly four prime factors, which may not be distinct.
\end{promptbox}

\begin{promptbox}{Solution generated by Gemini}
\small
Let the given expression be $f(n) = n^2 - n + 11$. We are looking for the smallest natural number $n$ such that $f(n)$ has exactly four prime factors, including multiplicity. This means $\Omega(f(n)) = 4$, where $\Omega(k)$ is the total number of prime factors of $k$.

Let's analyze the expression $f(n) = n(n-1) + 11$.
Since $n(n-1)$ is always an even number, $f(n)$ is always odd. So, 2 cannot be a prime factor of $f(n)$.
Let's check for other small prime factors:
- Modulo 3: $n(n-1)$ can be $0 \times (-1) \equiv 0$, $1 \times 0 \equiv 0$, or $2 \times 1 \equiv 2$. So $n^2-n+11 \equiv n(n-1)+2 \pmod 3$ can be $0+2=2$, $0+2=2$, or $2+2=4 \equiv 1$. So $f(n)$ is never divisible by 3.
- Modulo 5: $n(n-1)+11 \equiv n(n-1)+1 \pmod 5$. $n(n-1)$ can be $0, 2, 6\equiv 1, 12\equiv 2, 20\equiv 0$. So $n(n-1)+1 \pmod 5$ can be $1, 2, 3$. So $f(n)$ is never divisible by 5.
- Modulo 7: $n(n-1)+11 \equiv n(n-1)+4 \pmod 7$. $n(n-1)$ can be $0, 2, 6, 12\equiv 5, 20\equiv 6, 30\equiv 2, 42\equiv 0$. So $n(n-1)+4 \pmod 7$ can be $4, 6, 10\equiv 3, 9\equiv 2$. So $f(n)$ is never divisible by 7.
- Modulo 11: $f(n) = n^2-n+11 \equiv n(n-1) \pmod{11}$. $f(n)$ is divisible by 11 if and only if $n(n-1) \equiv 0 \pmod{11}$. Since 11 is prime, this occurs when $n \equiv 0 \pmod{11}$ or $n \equiv 1 \pmod{11}$.

So, the smallest possible prime factor of $f(n)$ is 11.
We want $f(n)$ to be a product of four primes, $p_1 p_2 p_3 p_4$. To find the smallest $n$, we should look for the smallest possible value of $f(n)$. The function $f(n)=n^2-n+11$ is increasing for $n \ge 1$. So smaller values of $n$ give smaller values of $f(n)$.

The smallest number that is a product of four primes, where each prime is at least 11, is $11^4$.
Let's see if $f(n)$ can be equal to $11^4$.
$11^4 = 14641$.
We need to solve $n^2 - n + 11 = 14641$ for a natural number $n$.
$n^2 - n - 14630 = 0$.
Using the quadratic formula, $n = \frac{-(-1) \pm \sqrt{(-1)^2 - 4(1)(-14630)}}{2(1)} = \frac{1 \pm \sqrt{1 + 58520}}{2} = \frac{1 \pm \sqrt{58521}}{2}$.
For $n$ to be an integer, 58521 must be a perfect square.
Let's estimate $\sqrt{58521}$. $200^2 = 40000$, $250^2 = 62500$.
$240^2 = 57600$. $241^2 = (240+1)^2 = 57600 + 480 + 1 = 58081$.
$242^2 = (240+2)^2 = 57600 + 960 + 4 = 58564$.
Since $241^2 < 58521 < 242^2$, 58521 is not a perfect square.
Thus, there is no integer $n$ for which $f(n) = 11^4$.

Let's find the next smallest number composed of four prime factors, each at least 11.
The candidates for the smallest such numbers are $11^4=14641$, $11^3 \times 13 = 1331 \times 13 = 17303$, $11^2 \times 13^2 = 121 \times 169 = 20449$, etc.
The smallest is $11^4$, which we've ruled out. The next smallest is $11^3 \times 13$.
Let's check if $f(n)$ can be equal to $11^3 \times 13 = 17303$.
We need to solve $n^2 - n + 11 = 17303$.
$n^2 - n - 17292 = 0$.
Using the quadratic formula, $n = \frac{1 \pm \sqrt{1 - 4(1)(-17292)}}{2} = \frac{1 \pm \sqrt{1 + 69168}}{2} = \frac{1 \pm \sqrt{69169}}{2}$.
For $n$ to be an integer, 69169 must be a perfect square.
Let's estimate $\sqrt{69169}$. $200^2=40000$, $300^2=90000$.
$260^2 = 67600$. The last digit is 9, so the root could end in 3 or 7.

Let's try $263^2 = (260+3)^2 = 67600 + 2 \times 3 \times 260 + 9 = 67600 + 1560 + 9 = 69169$.
So, 69169 is a perfect square, and its square root is 263.
Then $n = \frac{1 \pm 263}{2}$.
This gives two possible integer values for $n$:
$n_1 = \frac{1+263}{2} = \frac{264}{2} = 132$.
$n_2 = \frac{1-263}{2} = \frac{-262}{2} = -131$.
Since we are looking for a natural number $n$, we have $n=132$.
For $n=132$, $f(132) = 132^2 - 132 + 11 = 17424 - 132 + 11 = 17292 + 11 = 17303$.
And $17303 = 11^3 \times 13$. The prime factors are 11, 11, 11, 13. There are exactly four prime factors.
So, $n=132$ is a candidate.

To be sure that $n=132$ is the smallest such natural number, we need to confirm that no smaller $n$ results in a four-prime-factor number.
The function $f(n)$ is increasing for $n \ge 1$. So for any $n < 132$, $f(n) < f(132) = 17303$.
If there were a smaller $n_0 < 132$ such that $f(n_0)$ has four prime factors, then $f(n_0)$ would have to be a number smaller than 17303 with four prime factors, each being at least 11.
The smallest such number is $11^4 = 14641$. We have already shown that $f(n)$ cannot be 14641.
The next smallest such number is $11^3 \times 13 = 17303$, but this value is not smaller than 17303. Any other combination of four primes (e.g., $11^2 \times 13^2 = 20449$) will be larger.
Therefore, there is no value of $f(n)$ with four prime factors smaller than 17303.
This means that no $n < 132$ can produce a value of $f(n)$ with four prime factors.
Thus, the smallest natural number $n$ is 132.

Final check:
For $n=132$, $f(132) = 132^2 - 132 + 11 = 17303$.
Prime factorization of $17303$:
$17303 = 11 \times 1573$
$1573 = 11 \times 143$
$143 = 11 \times 13$
So $f(132) = 11 \times 11 \times 11 \times 13 = 11^3 \times 13$.
This number has $3+1=4$ prime factors.

\end{promptbox}

\end{document}